\documentclass[journal]{IEEEtran}

\ifCLASSINFOpdf
\else
\fi

\usepackage{amsmath,amsthm}
\newtheorem{theorem}{Theorem}

\usepackage{mystyle}
\usepackage{booktabs} %
\usepackage{multirow}
\usepackage{graphicx}
\graphicspath{{./images/}}

\usepackage{resizegather}

\begin{document}
\title{Tensor Dropout for Robust Learning}

\author{Arinbjörn Kolbeinsson* \qquad \qquad
        Jean Kossaifi* \qquad \qquad
        Yannis Panagakis \qquad \qquad
        Adrian Bulat\\ \vspace{10pt}
        Animashree Anandkumar\qquad \qquad
        Ioanna Tzoulaki\qquad \qquad
        Paul M. Matthews
\thanks{* indicates equal contribution}
\thanks{A. Kolbeinsson, P. M. Matthews are with Imperial College London}%
\thanks{I. Tzoulaki is with Imperial College London \&  University of Ioannina}
\thanks{J. Kossaifi and A. Anandkumar are with NVIDIA}%
\thanks{Y. Panagakis is with University of Athens}%
\thanks{A. Bulat is with Samsung AI}%
}

\maketitle

\begin{abstract}
CNNs achieve remarkable performance by leveraging deep, over-parametrized architectures, trained on large datasets. However, they have limited generalization ability to data outside the training domain, and a lack of robustness to noise and adversarial attacks. By building better inductive biases, we can improve robustness and also obtain smaller networks that are more memory and computationally efficient. While standard CNNs use matrix computations, we study tensor layers that involve higher-order computations and provide better inductive bias. Specifically, we impose low-rank tensor structures on the weights of tensor regression layers to obtain compact networks, and propose tensor dropout, a randomization in the tensor rank for robustness.
We show that our approach outperforms other methods for large-scale image classification on ImageNet and CIFAR-100. We establish a new state-of-the-art accuracy for phenotypic trait prediction on the largest dataset of brain MRI, the UK Biobank brain MRI dataset, where multi-linear structure is paramount. In all cases, we demonstrate superior performance and significantly improved robustness, both to noisy inputs and to adversarial attacks. We rigorously validate the theoretical validity of our approach by establishing the link between our randomized decomposition and non-linear dropout.
\end{abstract}

\IEEEpeerreviewmaketitle

\section{Introduction}

Deep neural networks have evolved into powerful predictive models with remarkable performance on computer vision tasks~\cite{krizhevsky2012imagenet,lecun2015deep,he2016deep}. Such models are usually over-parameterized, involving an enormous number (typically tens of millions) of parameters. This is much larger than the typical number of available training samples, making deep networks prone to overfitting~\cite{caruana2001overfitting}. Coupled with overfitting, deep networks lack robustness to small adversarial or noise perturbations~\cite{goodfellow2014explaining}. In computer vision tasks, such as image classification or regression, perturbed examples are often perceived identically to the original ones by humans while lead arbitrarily different predictions by networks. In addition, not only are these neural networks lacking in robustness, they are also over-confident when (wrongly) predicting on noisy inputs~\cite{hein2019relu}.
These shortcomings pose a possible obstacle for mass deployment of systems relying on deep learning in sensitive fields such as medical image analysis for disease prediction and expose an inherent weakness in their reliability.

To improve the robustness of deep neural networks to (adversarial or random) perturbations
and prevent them from overfitting, several methods that essentially affect the parameters of deep networks via regularization have been investigated. In particular, the link between regularization and robustness of deep neural networks to adversarial perturbations have been recently established in~\cite{bietti2019kernel,jakubovitz2018improving}.
These regularization mostly fall into two categories: structural changes to the architecture to make them inherently more robust and regularization methods that constrain directly the parameters. 
The latter category is well studied in the context of deep learning. In this context, regularization can be applied in the form of added randomness during training and/or testing. This can be applied to the activations through dropout~\cite{srivastava2014dropout} or to the weights (DropConnect~\cite{wan2013regularization}). Regularization functions (e.g., $\ell_2$- or $\ell_1$- norm) can also be directly applied to network’s parameters~\cite{nowlan1992simplifying,krogh1992simple,scardapane2017group,zhang2016l1}.

However, a more powerful class of regularization focuses on structural changes to the architecture of the network itself. The most notable such approach is based on tensor methods and accounts for the structure in the data. Machine learning is in essence the learning from data, and tensor methods allow to fully leverage the structure in that data. This  is  because  the  unknown  transformation
from  input  image  to  the  output  label  is  in  general  a  tensor map.
The data typically manipulated by deep neural networks is typically 3 dimensional (e.g. RGB images, brain MRI) or 4 dimensional (e.g. videos, functional MRI images, etc). Preserving multi-dimensional structure is crucial for performance.  
By limiting the network to matrix operations (e.g. with flattening layers followed by one or more fully-connected), we are ignoring this structure, resulting in deteriorated performance~\cite{kossaifi2017tensor,kossaifi2018tensor,kossaifi2019efficient}.
In contrast, tensor methods allow to leverage that structure to improve the model, reduce the number of parameters and improve computational efficiency.  One way this is done is by leveraging the multi-linear correlations in the network~\cite{tai2015convolutional,cheng2015exploration,yu2017compressing,kossaifi2018tensor,kossaifi2019t}.

In this paper, we introduce a novel higher-order randomized factorization method to improve the performance and robustness of deep networks to adversarial and random perturbations. We apply this stochastic decomposition to tensor regression layers which we use to entirely replace flattening and fully-connected layers. This preserves the structure of multi-dimensional data (e.g., images and MRI data) by expressing an output tensor as the result of a tensor contraction between the input tensor and some low-rank regression weight tensors. Instead of using the deterministic reconstruction of the regression weight tensor, our randomized tensor decomposition leads to the stochastic reduction of the rank during both training and inference. This is related to non-linear dropout on tensor factorization, which, by randomly dropping units during training, prevents over-fitting. However, rather than dropping random elements from the \emph{activation} tensor, this is done on the rank of the regression weight tensor.

We conduct thorough experiments in image classification and phenotypic trait prediction from MRI analysis. Specifically, we apply our method to the UK Biobank brain MRI dataset for estimating brain-age, a metric which has been associated with a range of diseases and mortality \cite{cole2017brain}, and could be an early predictor for Alzheimer's disease \cite{franke2012brain}. A more accurate and robust metric of brain condition can consequently lead to more accurate disease diagnoses.

\textbf{Summary of contributions:}
\begin{itemize}%
\item We propose tensor dropout, a novel method, a novel stochastic tensor decomposition where non-linear dropout is applied in the latent subspace spanned by a low-rank factorization.
\item We apply tensor dropout to tensor regression layers and show that it improves the inductive bias of convolutional neural networks (CNNs) by fully leveraging the structure in the data via a stochastic tensor decomposition.
\item We demonstrate superior performance against both regular deep network architectures with fully-connected layers and networks with tensor regression layers that do not incorporate our proposed randomized decomposition.
\item We establish a new state-of-the-art for large scale regression from MRI data on the UK Biobank dataset, the largest dataset of brain MRI. We demonstrate that our model is significantly more robust to noise in the input, as occurs naturally during capture.
\item Our method makes neural networks significantly more robust to adversarial noise, \emph{without} adversarial training.
\item We show that our proposed method implicitly regularizes the tensor decomposition. We establish theoretically and empirically the link between tensor dropout and the deterministic low-rank tensor regression.
\item We validate all aspects of our method on the CIFAR-100 image classification dataset in thorough ablation studies.
\end{itemize}

\section{Related work}
In this section, we review the most closely related work to ours.

\textbf{Network regularization and dropout}. Several methods that improve generalization by mitigating overfitting have been developed in the context of deep learning. The interested reader is referred to the work of \cite{kukavcka2017regularization} and the references therein for a comprehensive survey of regularization techniques for deep networks.
The most closely related regularization method to our approach is Dropout~\cite{srivastava2014dropout}, which is probably the most widely adopted technique for training neural networks while preventing overfitting. Concretely, during training, each unit (i.e., neuron) is equipped with a binary Bernoulli random variable and only the network’s weights whose corresponding Bernoulli variables are sampled with value \(1\) are updated at each back-propagation step. At each iteration, those Bernoulli variables are re-sampled again and the weights are updated accordingly. Our proposed regularization method can be interpreted as dropout on low-rank tensor regression, a fact which we proved in Section~(\ref{ssec:trl-bernouilli}).

\textbf{Randomized tensor decompositions}.
Tensor decompositions exhibit high computational cost and low convergence rate when applied to massive multi-dimensional data. To accelerate computation, and enable them to scale, randomized approaches have been employed. A randomized least squares algorithm for CP decomposition is proposed by~\cite{battaglino2018practical}, which is significantly faster than traditional one. In~\cite{erichson2017randomized}, CP is applied on a small tensor generated by multi-linear random projection of the high-dimensional tensor. The CP decomposition of the large-scale tensor is obtained by back projection of the CP decomposition of the small tensor. \cite{wang2015fast} introduce a fast yet provable randomized CP decomposition that performs randomized tensor contraction using FFT. Methods in~\cite{sidiropoulos2014parallel,vervliet2014breaking} are highly computationally efficient algorithms for computing large-scale CP decompositions by applying random projections into a set of small tensors, derived by subdividing a tensor into a set of blocks. Fast randomized algorithms that approximate Tucker decomposition using Sketching have also been investigated \cite{tsourakakis2010mach,zhou2014decomposition}. More recently, a randomized tensor ring decomposition that employs tensor random projections has been developed in~\cite{yuan2019randomized}. The most similar method to ours is that of \cite{battaglino2018practical}, where elements of the tensor are sampled randomly, and each factor of the decomposition updated in an iterative manner. By contrast, our method allows for end-to-end training, and applies randomization on the \emph{fibers} of the tensor, effectively randomizing the rank of the weight tensor.

\section{Preliminaries and tensor regression layers}
In this section, we introduce the notations and preliminary notions necessary to introduce our stochastic rank regularization, including tensor regression layers.

\begin{figure*}
    \begin{subfigure}[t]{0.45\textwidth}
        \centering
        \includegraphics[height=3cm]{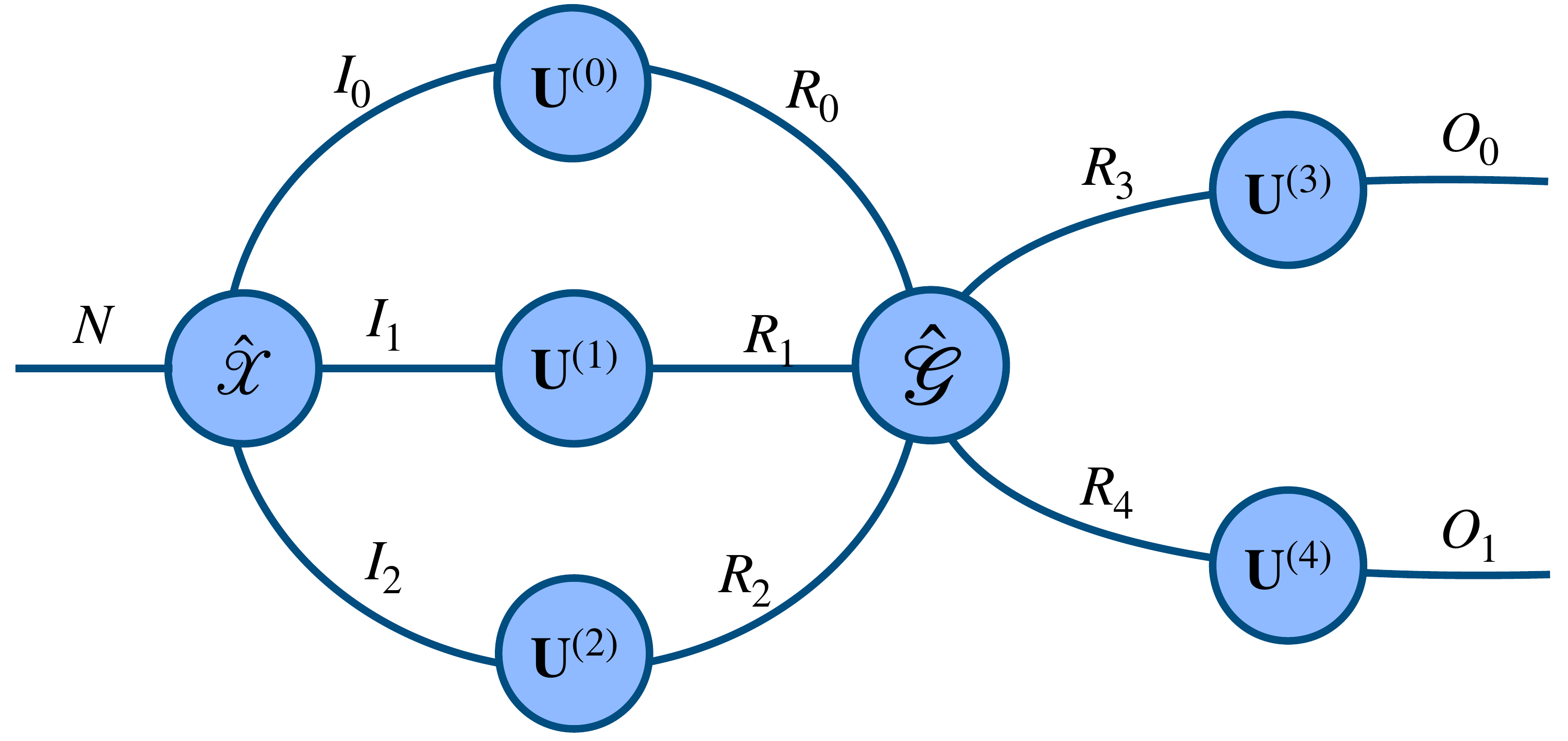}
        \caption{Tensor diagram of a TRL}
        \label{fig:trl-diagram}
    \end{subfigure}
    \qquad
    \begin{subfigure}[t]{0.45\textwidth}
        \centering
        \includegraphics[height=2.9cm]{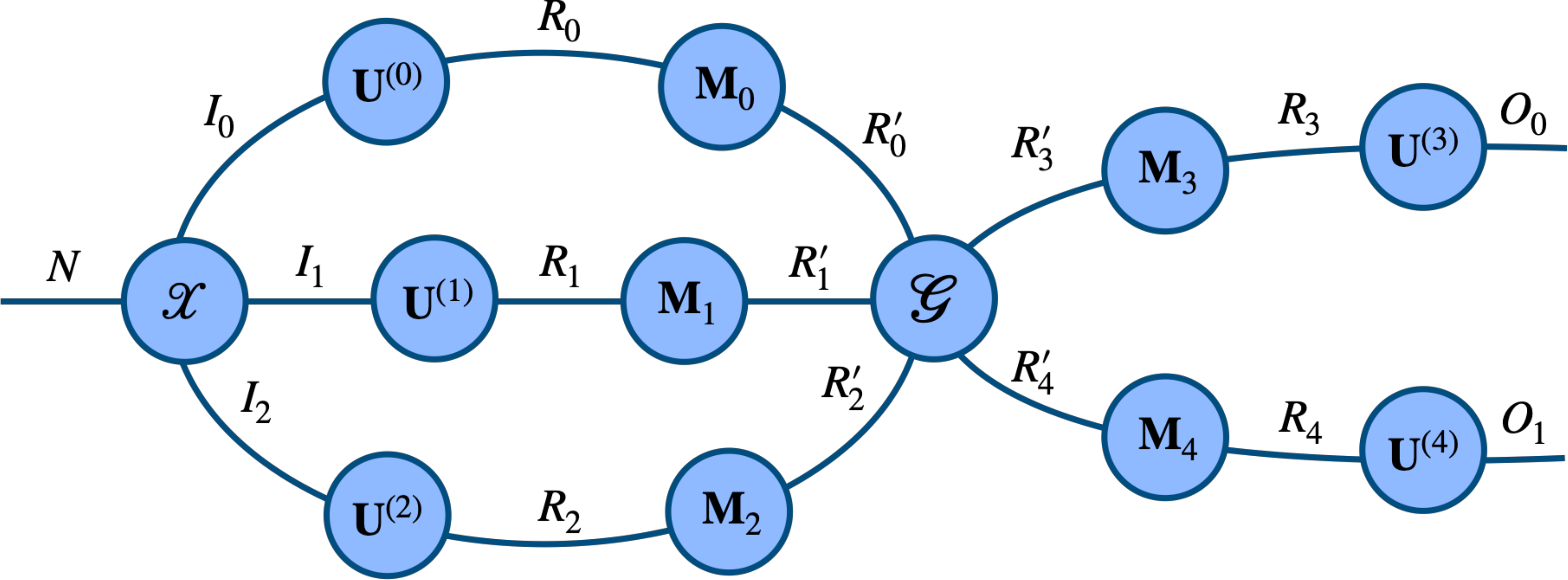}
        \caption{Tensor diagram of a R-TRL}
        \label{fig:srr-trl-diagram}
    \end{subfigure}
    \caption{Tensor diagrams of the TRL (left) and our proposed R-TRL (right), with low-rank constraints imposed on the regression weights tensor using a Tucker decomposition. Note that the CP case is readily given by this formulation by additionally having the core tensor \mytensor{G} be super-diagonal, and setting \( \mymatrix{M} = \mymatrix{M}^{(0)} = \cdots = \mymatrix{M}^{(N)} = \mydiag(\myvector{\lambda}).\)}
    \label{fig:trl-diagrams}
\end{figure*}

\textbf{Notation:} We denote \(\myvector{v}\) to be vectors (1\myst order tensors) and \(\mymatrix{M}\) to be matrices (2\mynd order tensors). \(\myId\) is the identity matrix. We denote as \(\mytensor{X}\) tensors of order \(N \geq 3\), and denote its element \((i, j, k)\) as \(\mytensor{X}_{i_0, i_1, \cdots, i_{N -}} \,\text{ or } \, \mytensor{X}(i_0, i_1, \cdots, i_{N -})\). A colon is used to denote all elements of a mode e.g. the mode-0 fibers of \(\mytensor{X}\) are denoted as \(\mytensor{X}_{\mycolon, i_1, \cdots, i_{N}}\). The transpose of \(\mymatrix{M}\) is denoted \(\mymatrix{M}\myT\). %
For any tensor \(\mytensor{X}\), we denote \(\mytensor{X}^{*2}\) the Hadamard power of 2, which squares each element of the tensor.
Finally, for any \(i, j \in \myN, i < j,\, \myrange{i}{j}\) denotes the set of integers \(\{ i, i+1, \cdots , j-1, j\}\), and \(i \mydiv j\) the integer division of \(i\) by \(j\).

\textbf{Tensor unfolding:}
	Given a tensor,
	\( \mytensor{X} \in \myR^{I_0 \times I_1 \times \cdots \times I_{N}}\),
	its mode-\(n\) unfolding is a matrix \(\mymatrix{X}_{[n]} \in \myR^{I_n, I_M}\), 
	with \(M = \prod_{\substack{k=0,\\k \neq n}}^{N} I_k\), defined by the mapping from 
	\( (i_0, i_1, \cdots, i_N)\) to \((i_n, j)\), with 
	\[
	j = \sum_{\substack{k=0,\\k \neq n}}^{N} i_k \times \prod_{\substack{m=k+1,\\ m \neq n}}^{N} I_m.
	\]
	
\textbf{Tensor vectorization:} 	Given a tensor,
	\( \mytensor{X} \in \myR^{I_0 \times I_1 \times \cdots \times I_{N}}\), 
	we can flatten it into a vector \(\text{vec}(\mytensor{X})\) 
	of size \(\left(I_0 \times \cdots \times I_{N}\right)\) 
	defined by the mapping from element
	\( (i_0, i_1, \cdots, i_{N})\) of \(\mytensor{X}\) to element \(j\) of \(\text{vec}(\mytensor{X})\), with 
	\[ j = \sum_{k=0}^{N} i_k \times \prod_{m=k+1}^{N} I_m.\]

\textbf{Mode-n product:}
For a tensor \(\mytensor{X} \in \myR^{I_0 \times I_1 \times \cdots \times I_{N}}\) and a matrix \( \mymatrix{M} \in \myR^{J \times I_n} \), the n-mode product of a tensor 
is a tensor of size 
\(\left(I_0 \times \cdots \times I_{n-1} \times J \times I_{n+1} \times \cdots \times I_{N}\right)\) 
and can be expressed using unfolding of \(\mytensor{X}\) 
and the classical dot product as:
\[
	\left(\mytensor{X} \times_n \mymatrix{M}\right)_{[n]} = \mymatrix{M} \mytensor{X}_{[n]} \in \myR^{I_0 \times \cdots \times I_{n-1} \times R \times I_{n+1} \times \cdot \times I_{N}}
\]

\textbf{Inner product:}
For two tensors of matching sizes, \(\mytensor{X}, \mytensor{Y} \in \myR^{I_0  \times \cdots \times I_{N}}\), we denote by 
\(\myinner{\mytensor{X}}{\mytensor{Y}} \in \myR\) the contraction of \(\mytensor{X} \) by \(\mytensor{Y}\) along each mode.
\[
	\myinner{\mytensor{X}}{\mytensor{Y}}_N = \sum_{i_0=0}^{I_0}\sum_{i_1=0}^{I_1} \cdots \sum_{i_n=0}^{I_{N}} \mytensor{X}_{i_0, i_1, \cdots, i_n} \mytensor{Y}_{i_0, i_1, \cdots, i_n}
\]

\textbf{Kruskal tensor:}
Given a tensor \(\mytensor{X} \in \myR^{I_0 \times I_1 \times \cdots \times I_{N - 1}} \), 
the Canonical-Polyadic decomposition (CP), also called PARAFAC, decomposes it into a sum of \(R\) rank-1 tensors. The number of terms in the sum, \(R\), is known as the rank of the decomposition. Formally, we find the vectors \( \mathbf{u}^{(0)}_{k} , \mathbf{u}^{(1)}_{k}, \cdots, \mathbf{u}^{(N )}_{k}\), for \(k = \myrange{0}{R-1}\), as well as an optional vector of weights \(\myvector{\lambda} \in \myR^R\) to absorbs the magnitude of the factors such that:
\begin{equation}
\mytensor{X}
= \sum_{k=0}^{R-1}  \underbrace{ \lambda_k \myvector{u}^{(0)}_{k} \circ \myvector{u}^{(1)}_{k} \circ \cdots \circ \myvector{u}^{({N})}_{k} }_{\text{rank-1 components}},
\end{equation}

For any \(k \in \myrange{0}{{N}},\) these vectors \(\mathbf{u}^{(k)}_{r}, r \in \myrange{0}{R-1}\) can be collected in matrices, called factors or the decomposition, and defined as \(\mathbf{U}^{(k)} = \left[ \begin{matrix} \mathbf{u}^{(k)}_{0}, \mathbf{u}^{(k)}_{1}, \cdots, \mathbf{u}^{(k)}_{R-1} \end{matrix} \right].\)
The decomposition can be denoted more compactly as
\(
\mytensor{X} = \mykruskal{\myvector{\lambda};\, \mathbf{U}^{(0)}, \cdots, \mathbf{U}^{({N })}}
\).%

\textbf{Tucker tensor:}
Given a tensor \(\mytensor{X} \in \myR^{I_0 \times I_1 \times \cdots \times I_{N}} \), 
we can decompose it into a low rank core 
\(\mytensor{G} \in \myR^{R_0 \times R_1 \times \cdots \times R_{N -}}\) 
by projecting along each of its modes
with projection factors 
\( \left( \mymatrix{U}^{(0)}, \cdots,\mymatrix{U}^{({N})} \right) \), with \(\mymatrix{U}^{(k)} \in \myR^{R_k, I_k}, k \in (0, \cdots, {N})\).
The tensor in its decomposed form  can then be written:
\begin{align}
\mytensor{X} &= 
\mytensor{G} \times_0 \mymatrix{U}^{(0)} 
		  \times_1  \mymatrix{U}^{(1)} \times
		  \cdots
          \times_{N} \mymatrix{U}^{({N})} \nonumber \\
        &= \mytucker{\mytensor{G}}{\mymatrix{U}^{(0)},
		  \cdots,
          \mymatrix{U}^{({N})}} 
\end{align}
Note that the Kruskal form of a tensor can be seen as a Tucker tensor with a super-diagonal core.

\textbf{Tensor diagrams:} In order to represent easily tensor operations, we adopt the tensor diagrams, where tensors are represented by vertices (circles) and edges represent their modes. The degree of a vertex then represents its order. Connecting two edges symbolizes a tensor contraction between the two represented modes. Figure~\ref{fig:trl-diagrams} presents a tensor diagram of the tensor regression layer and its stochastic rank-regularized counter-part.

 \textbf{Tensor regression layers:}
 Reducing the storage and computational costs of deep networks has become critical for meeting the
requirements of environments with limited memory or computational resources.
To this end, a surge of network compression and approximation algorithms have recently been proposed in the context of deep learning. By leveraging the redundancy in network parameters, methods such as \cite{tai2015convolutional,cheng2015exploration,yu2017compressing,kossaifi2018tensor} employ low-rank approximations
of deep networks’ weight matrices (or tensors) for parameter reduction. Network compression methods in the frequency domain \cite{chen2016compressing} have also been investigated. An alternative approach for reducing the number of effective parameters in deep nets relies on sketching, whereby, given a matrix or tensor of input data or parameters, one first compresses it to a much smaller matrix (or tensor) by multiplying it by a (usually) random matrix with certain properties \cite{kasiviswanathan2017deep,daniely2016sketching}.

A particularly appealing approach to network compression, especially for visual data (and other types of multidimensional and multi-aspect data), is tensor regression networks~\cite{kossaifi2018tensor}. Deep neural networks typically leverage the spatial structure of input data via series of convolutions, point-wise non-linearities, pooling, etc.
However, this structure is usually wasted by the addition, at the end of the networks' architectures, of a flattening layer followed by one or several fully-connected layers. 
A recent line of study focuses on alleviating this using tensor methods. \cite{kossaifi2017tensor} proposed tensor contraction as a layer, to reduce the size of activation tensors, and demonstrated large space savings with this layer. 
Recently, tensor regression networks \cite{kossaifi2018tensor} propose to replace fully-connected layers entirely with a tensor regression layer (TRL). This tensor regression layers preserves the structure by expressing an output tensor as the result of a tensor contraction between the input tensor and some low-rank regression weight tensors.

More specifically, we assume we have a set of \(S\) input activation tensors
\(\mytensor{X}^{(k)} \in \myR^{I_0 \times I_1 \times \cdots \times I_{N}} \)
and corresponding target labels \( y^{(k)} \in \myR \), with \(k \in \myrange{0}{N}\). Not that here, for clarity, we introduce the tensor regression layer for scalar targets. In practice the target is typically a vector or a tensor.
A tensor regression layer estimates the regression weight tensor 
\( \mytensor{W} \in 
   \myR^{I_0 \times I_1 \times \cdots \times I_N},
\)
expressed under some low-rank decomposition.
For instance, using a rank--\(\left(R_0, \cdots, R_N\right)\) Tucker decomposition, we have:
\begin{align}\label{eq:trl}
    y^{(k)} & = \myinner{\mytensor{X}^{(k)}}{\mytensor{W}} + b \nonumber \\
    \text{with } & 
         \mytensor{W} 
         =  
         \mytensor{G} 
         \times_0 \mymatrix{U}^{(0)}
         \times_1 \mymatrix{U}^{(1)}
         \cdots
         \times_N \mymatrix{U}^{(N)}
\end{align}
with \(\mytensor{G} \in \myR^{R_0 \times \cdots \times R_N} \), 
\(\mymatrix{U}^{(k)} \in \myR^{I_k \times R_k}\) 
for each 
\(k\) in \(\myrange{0}{N}\)
and \(\mymatrix{U}^{(N)} \in \myR^{O \times R_{N}}\).

In addition to preserving and leveraging the structure in the input, TRL allow for large space savings without sacrificing accuracy. \cite{cao2017tensor} explore the same model with various low-rank structures on the regression weights.

\section{Tensor dropout}\label{ssec:trl-bernouilli}

In this section, we introduce our tensor dropout and apply it to tensor regression layers. Specifically, we propose a new stochastic rank-regularization, applied to low-rank tensors in decomposed forms. This formulation is general and can be applied to any type of decomposition. We introduce it here, without loss of generality, to the case of Tucker and CP decompositions.

\textbf{Randomized tensor regression layer}
We propose a novel randomized decomposition on \(\mytensor{W}\), which applies dropout in the latent subspace spanned by a tensor decomposition.
For instance, for an \(N\myth\)--order regression weight with a Tucker structure, we can define for each \(k \in \myrange{0}{N}\), a sketch matrix \(\mymatrix{M}^{(k)} \in \myR^{R_{k} \times R_{k}}\) (e.g. a random projection or column selection matrix). This can then be used to sketch the factors \(\mymatrix{U}^{(k)}\) of the decomposition as \(\mymatrix{\tilde U}^{(k)} = \mymatrix{U}^{(k)}(\mymatrix{M}^{(k)})\myT\) be a sketch of factor matrix , and the core tensor  \(\mytensor{G}\) as \(\mytensor{\tilde G} = \mytensor{G} \times_0  \mymatrix{M}^{(0)} \times \cdots \times_{N} \mymatrix{M}^{(N)}\).

Applying this to tensor regression, we can apply this tensor dropout technique to the regression weights. Given an activation tensor \(\mytensor{X} \in \myR^{I_0 \times \cdots \times I_{N}}\) and a set of \(S\) target labels \(\myvector{y}^{(k)}\), a Randomized-Tensor Regression Layer (R-TRL) is obtained from equation~\ref{eq:trl} and aims at minimizing the empirical risk:
\begin{equation}\label{eq:rnd-trl}
\frac{1}{S-1}\sum^{S-1}_{k=0}
\left(\myvector{y}^{(k)} - \myinner{\mytensor{\tilde W}}{\mytensor{X}^{(k)}}
\right)^2,
\end{equation}
where \( \mytensor{\tilde W} \) is a stochastic low-rank approximation of Tucker decomposition. In other words, in addition to the low-rank structure of the weights, we apply our tensor dropout in the latent subspace spanned by the decomposition.

For instance, in the case of a Tucker R-TRL, we have:
\begin{equation}\label{eq:rnd-weight}
 \mytensor{\tilde W} 
 =  
 \mytensor{\tilde G}
 \times_0 \mymatrix{\tilde U}^{(0)} 
 \times
 \cdots
 \times_{N} \mymatrix{\tilde U}^{(N)}
\end{equation}

This is the core of our proposed R-TRL, which incorporates tensor dropout within a TRL. 
Even though several sketching methods have been proposed, we focus here on R-TRL with two different types of binary sketching matrices, namely binary matrix sketching with replacement and binary diagonal matrix sketching with Bernoulli entries, which we detail below.

\subsection{Bernoulli Tucker randomized tensor regression}
For any \(n \in \myrange{0}{N}\), let \(\myvector{\lambda}^{(n)}  \in \myR^{R_N}\) be a random vector, the entries of which are i.i.d. Bernoulli(\(\theta\)), then a diagonal Bernoulli sketching matrix is defined as \( \mymatrix{M}^{(n)} = \text{diag}(\myvector{\lambda}^{(n)})\).

When the low-rank structure on the weight tensor \(\mytensor{\tilde W}\) of the TRL is imposed using a Tucker decomposition, the randomized Tucker approximation is expressed as:
\begin{equation}\label{eq:bernouilli-tucker-weight}
\begin{split}
    \mytensor{\tilde W} 
    = & \,  
    \mytensor{G} \times_0  \mymatrix{M}^{(0)} \times \cdots \times_{N} \mymatrix{M}^{(N)}\\
    &
    \times_0 \left(\mymatrix{U}^{(0)}(\mymatrix{M}^{(0)})\myT \right)
    \times
    \cdots
    \times_{N} \left(\mymatrix{U}^{(N)}(\mymatrix{M}^{(N)})\myT \right) \\
    &  = \mytucker{\mytensor{\tilde G}}{\mymatrix{\tilde U}^{(0)}, \cdots, \mymatrix{\tilde U}^{(N)}}
\end{split}
\end{equation}

The main advantage of considering the above-mentioned sampling matrices is that the products \(\mymatrix{\tilde U}^{(k)} = \mymatrix{U}^{(k)}(\mymatrix{M}^{(k)})\myT\) or \(\mytensor{\tilde G} = \mytensor{G} \times_0  \mymatrix{M}^{(0)} \times \cdots \times_{N} \mymatrix{M}^{(N)}\) are never explicitly computed, we simply select the elements from \( \mytensor{ G}\) and the corresponding factors.

Interestingly, in analogy to dropout, where each hidden unit is dropped independently with probability \(1 - \theta\), in the proposed randomized tensor decomposition, the columns of the factor matrices and the corresponding fibers of the core tensor are dropped independently and consequently the \emph{rank} of the tensor decomposition is stochastically dropped.

The tensor dropout acts as an implicit regularizer on the regression, by limiting the rank, at each iteration. This can be shown by examining the expectation of the stochastic loss, which can be expressed deterministically as the unrandomized empirical loss, plus an additional regularization term.

\begin{theorem}{Tensor Dropout with Tucker decomposition is a deterministic regularized loss. (Proof in Appendix~\ref{sec:minimisation-proof})}\label{thm:tucker}

The minimization objective (equation~\ref{eq:rnd-trl}) can be reformulated by expanding the tensor contractions, and the expectation of the minimization objective becomes:
\begin{equation}\label{eq:rnd-trl-tucker}
\begin{split}
&\myexpectation{\myvector{\lambda}}{\frac{1}{S-1}\sum^{S-1}_{k=0}
\left(\myvector{y}^{(k)} - \myinner{\mytensor{\tilde W}}{\mytensor{X}^{(k)}}
\right)^2}\\
& = \frac{1}{S-1}\sum^{S-1}_{k=0}
\left(\myvector{y}^{(k)} - \theta^{N}\myinner{\mytensor{W}}{\mytensor{X}^{(k)}}
\right)^2 \\
& + \tiny{\frac{\theta^{N}(1-\theta^{N})}{S-1} \sum^{S-1}_{k=0}}
\myinner{\mytensor{G}^{\star 2}
\times_0 (\mymatrix{U}^{(0)})^{\star 2} \dots
\times_{N} (\mymatrix{U}^{(N)})^{\star 2}}{(\mytensor{X}^{(k)})^{\star 2}}.
\end{split}
\end{equation}

\end{theorem}

\subsection{Bernoulli CP randomized tensor regression}

An interesting special case of equation~(\ref{eq:rnd-weight}) is when the weight tensor \(\mytensor{\tilde W}\) of the TRL is expressed using a CP decomposition. 
In that case, we set \( \mymatrix{M} = \mymatrix{M}^{(0)} = \cdots = \mymatrix{M}^{(N)} = \mydiag(\myvector{\lambda}) \), with, for any \(k \in \myrange{0}{R}\), \(\lambda_k \sim \text{Bernoulli}(\theta).\)

Then a randomized CP approximation is expressed as:
\begin{equation}\label{eq:bernouilli-rnd-weight-long}
\begin{split}
    \mytensor{\tilde W} 
    & = 
    \sum_{k=0}^{R-1} \mymatrix{\tilde U}^{(0)}_k \circ \cdots \circ \mymatrix{\tilde U}^{(N)}_k
\end{split}
\end{equation}

The above randomized CP decomposition on the weights is equivalent to the following formulation (see proof in Appendix \ref{sec:bernoulli-proof}):
\begin{equation}\label{eq:bernouilli-rnd-weight}
\begin{split}
    \mytensor{\tilde W} 
    = &  
    \mykruskal{\myvector{\lambda};\, \mathbf{U}^{(0)}, \cdots, \mathbf{U}^{(N)}}
 \end{split}
\end{equation}

Based on the previous stochastic regularization, for an activation tensor \mytensor{X} and a corresponding label vector \(\myvector{y}\), the optimization problem for our tensor regression layer with stochastic regularization is given by:
\begin{equation}\label{eq:stochastic-problem}
    \min_{\mymatrix{U}^{(0)}, \cdots,  \mymatrix{U}^{(N)}} 
    \left(\myvector{y}^{(k)} - 
    \myinner{\mykruskal{\myvector{\lambda};\, \mathbf{U}^{(0)}, \cdots, \mathbf{U}^{(N)}}}{\mytensor{X}^{(k)}}
    \right)^2
\end{equation}

In addition, the above stochastic optimization loss can be rewritten as a deterministic regularized one:
\begin{theorem}{Tensor Dropout with CP decomposition is equivalent to a deterministic regularized loss. (Proof in Appendix~\ref{sec:bernoulli-proof})}\label{thm:CPminimisation}
\begin{equation}\label{eq:CPminimisation}
\begin{split}
&\myexpectation{\myvector{\lambda}}{\frac{1}{S-1}\sum^{S-1}_{k=0}
\left(\myvector{y}^{(k)} - 
    \myinner{\mykruskal{\myvector{\lambda};\, \mathbf{U}^{(0)}, \cdots, \mathbf{U}^{(N)}}}{\mytensor{X}^{(k)}}
\right)^2}\\
& = \frac{1}{S-1}\sum^{S-1}_{k=0}
\left(\myvector{y}^{(k)} - 
\theta\myinner{\mykruskal{\mathbf{U}^{(0)}, \cdots, \mathbf{U}^{(N)}}}{\mytensor{X}^{(k)}}
\right)^2 \\
& + \tiny{\frac{\theta(1-\theta)}{S-1} \sum^{S-1}_{k=0}}
\myinner{\mykruskal{(\mathbf{U}^{(0)})^{\star 2}, \cdots, (\mathbf{U}^{(N)})^{\star 2}}}{(\mytensor{X}^{(k)})^{\star 2}}.
\end{split}
\end{equation}
\end{theorem}

\subsection{R-TRL with replacement}\label{sec:replacement-rtrl}
Previously, we focus on the R-TRL with Bernoulli sampling only. Our model is more general and can be applied to many different sampling settings. 
Here, we introduce one such case: the R-TRL with binary sketching matrix (i.e. with replacement). Specifically, we first choose \(\theta \in [0, 1]\). 

Mathematically, we then introduce the uniform sampling matrices
\(
\mymatrix{M}^{(0)} \in \myR^{R_{0} \times R_{0}},
\cdots, 
\mymatrix{M}^{(N)} \in \myR^{R_{N} \times R_{N}}
\). 
\(\mymatrix{M}_j\) is a uniform sampling matrix, selecting \(K_j\) elements, where \(K_j = R_j \mydiv \theta\).
In other words, for any \(i \in \myrange{0}{N}\), \(\mymatrix{M}^{(i)}\) verifies:
\begin{equation}\label{eq:sampling-matrix}
    \mymatrix{M}^{(i)}(j, :) = 
    \begin{cases}
        \myvector{0} 
            & \mbox{ if } j > K \\ 
        \myId_m(r, :), m \in \myrange{0}{R_i} 
            & \mbox{ otherwise }
    \end{cases}
\end{equation}
In practice this is done efficiently by selecting directly the correct elements from \(\mytensor{G}\) and its corresponding factors.

\section{Experimental evaluation}

In this section, we present the numerical simulations, experimental setting, databases used and implementation details. We experimented on several datasets across various tasks, namely image classification and MRI-based regression.
All methods were implemented~\footnote{code will be released upon acceptance to reproduce all results} using PyTorch~\cite{paszke2017automatic} and TensorLy~\cite{kossaifi2019tensorly}. For all adversarial attacks, we used Foolbox~\cite{rauber2017foolbox}.

\subsection{Numerical experiments on synthetic data}\label{sec:synthetic}

\begin{figure*}[!ht]
    \centering
        \subcaptionbox*{}{
        \raisebox{50pt}{
            \rotatebox[origin=t]{90}{Objective}
        }
    }
    \begin{subfigure}[t]{0.18\textwidth}
        \centering
        \includegraphics[width=1\linewidth]{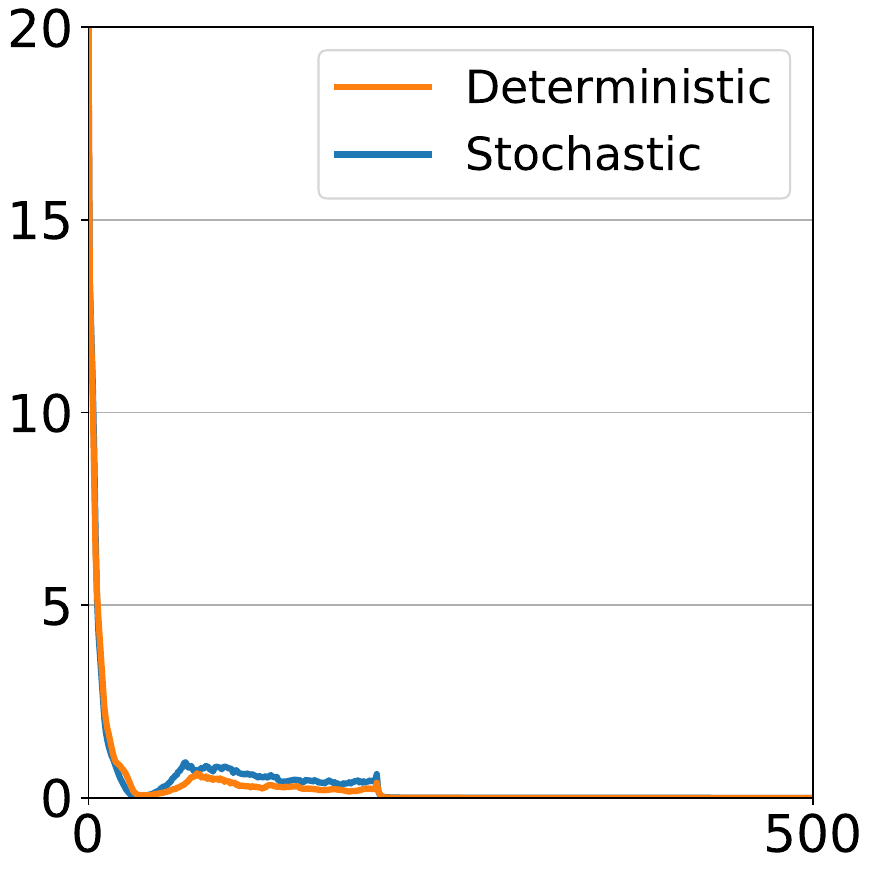}
        \caption{\textbf{\(\theta = 1\)}}
        \label{fig:synthetic-3-2}
    \end{subfigure}
    \begin{subfigure}[t]{0.18\textwidth}
        \centering
        \includegraphics[width=1\linewidth]{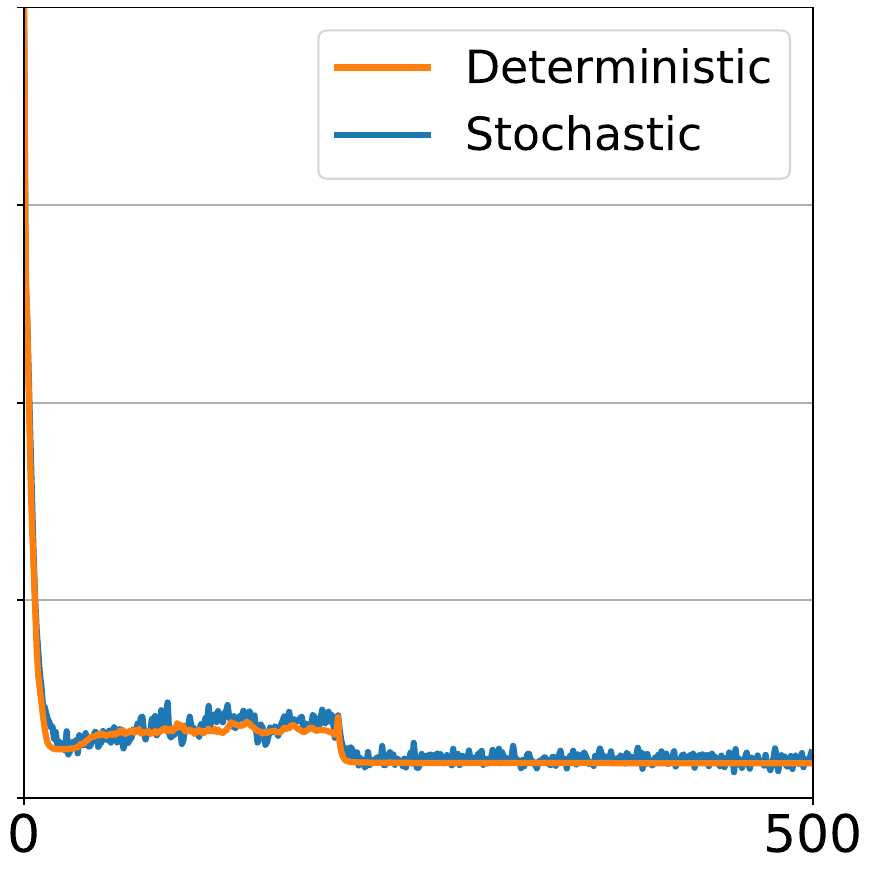}
        \caption{\textbf{\(\theta = 0.9\)}}
        \label{fig:synthetic-3-3}
    \end{subfigure}
    \begin{subfigure}[t]{0.18\textwidth}
        \centering
        \includegraphics[width=1\linewidth]{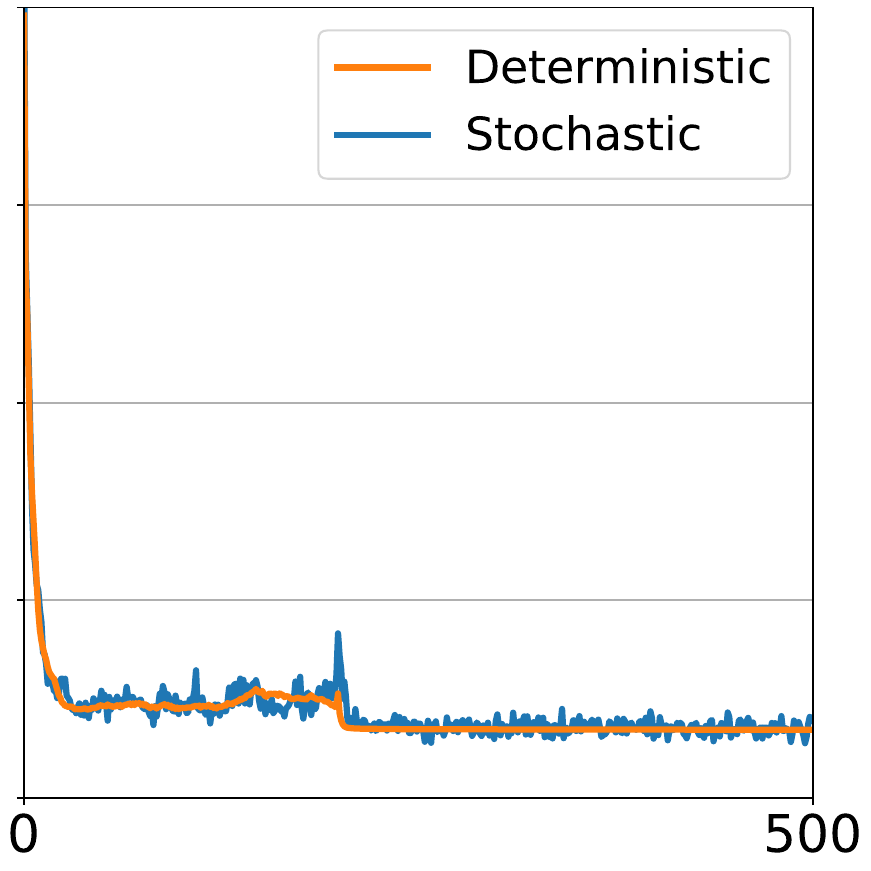}
        \caption{\textbf{\(\theta = 0.8\)}}
        \label{fig:synthetic-3-6}
    \end{subfigure}
    \begin{subfigure}[t]{0.18\textwidth}
        \centering
        \includegraphics[width=1\linewidth]{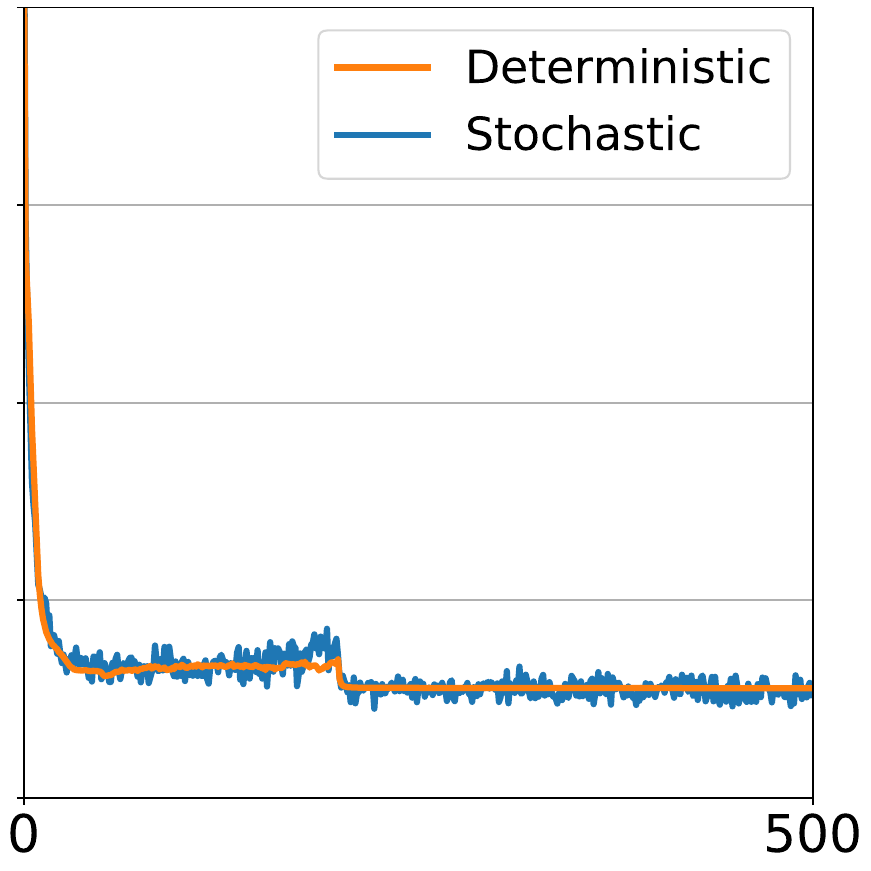}
        \caption{\textbf{\(\theta = 0.7\)}}
        \label{fig:synthetic-3-9}
    \end{subfigure}
    \begin{subfigure}[t]{0.18\textwidth}
        \centering
        \includegraphics[width=1\linewidth]{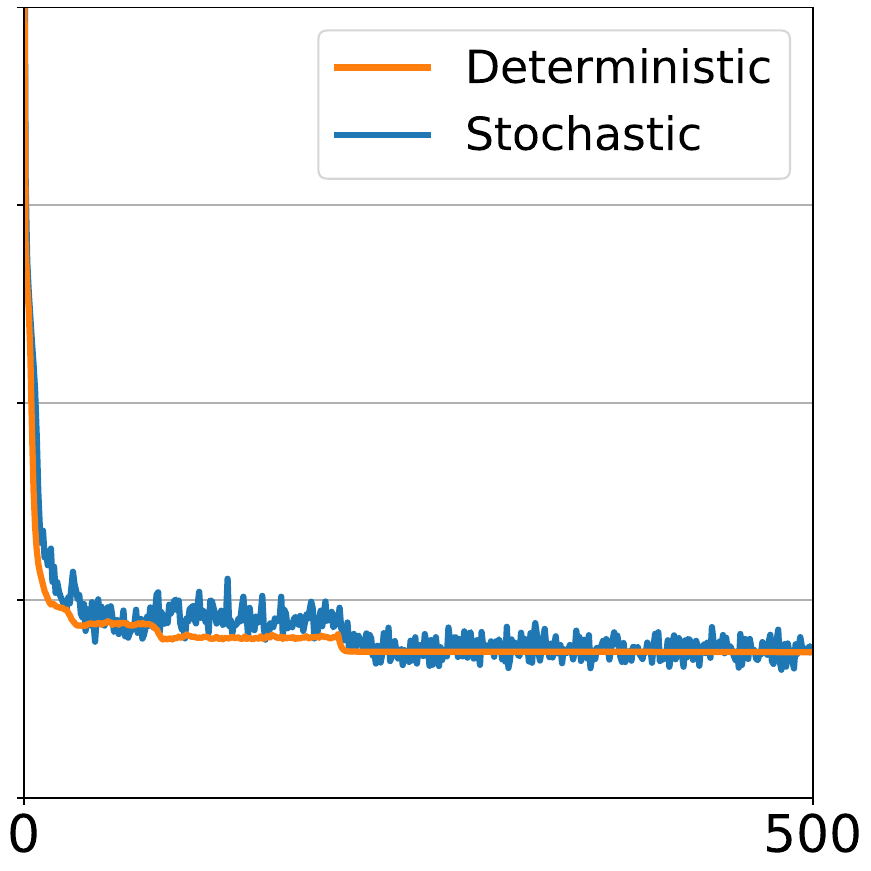}
        \caption{\textbf{\(\theta = 0.6\)}}
        \label{fig:synthetic-3-9}
    \end{subfigure}

    \caption{{\small\textbf{Experiment on synthetic data:} loss of the CP R-TRL as a function of the number of epochs for the stochastic version (orange) and the deterministic one based on the regularized objective function (blue). As expected, both formulations are empirically the same.}}
    \label{fig:synthetic}
\end{figure*}

\begin{figure*}[!ht]
    \centering
        \subcaptionbox*{}{
        \raisebox{50pt}{
            \rotatebox[origin=t]{90}{Objective}
        }
    }
    \begin{subfigure}[t]{0.18\textwidth}
        \centering
        \includegraphics[width=1\linewidth]{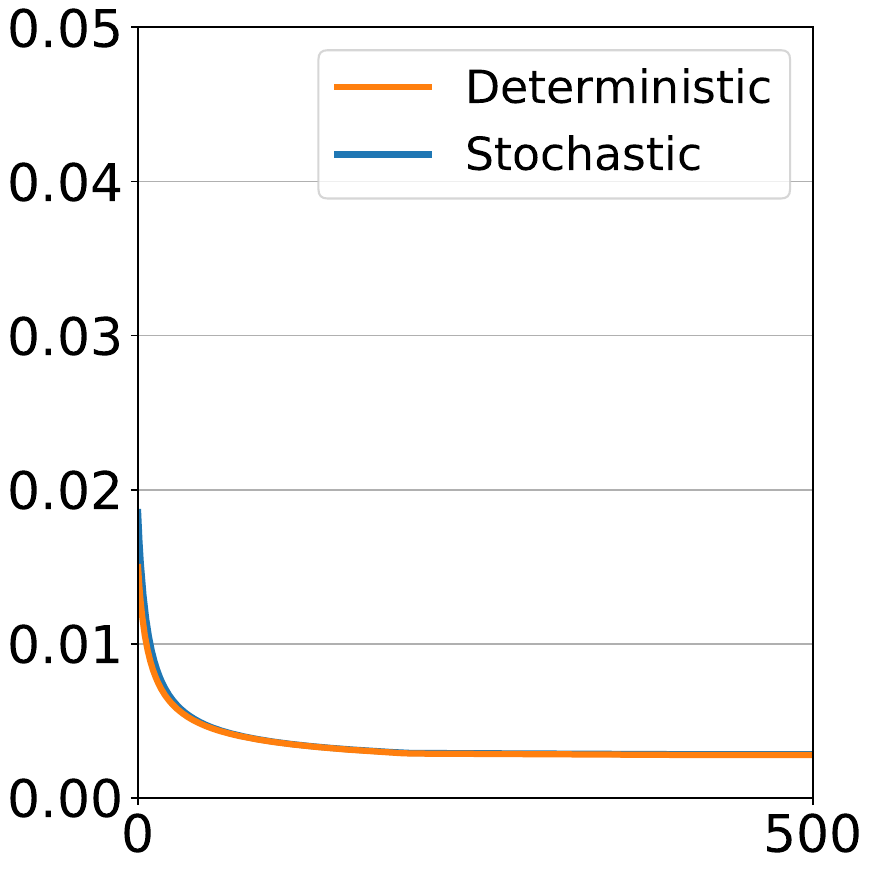}
        \caption{\textbf{\(\theta = 1\)}}
        \label{fig:synthetic-3-2}
    \end{subfigure}
    \begin{subfigure}[t]{0.18\textwidth}
        \centering
        \includegraphics[width=1\linewidth]{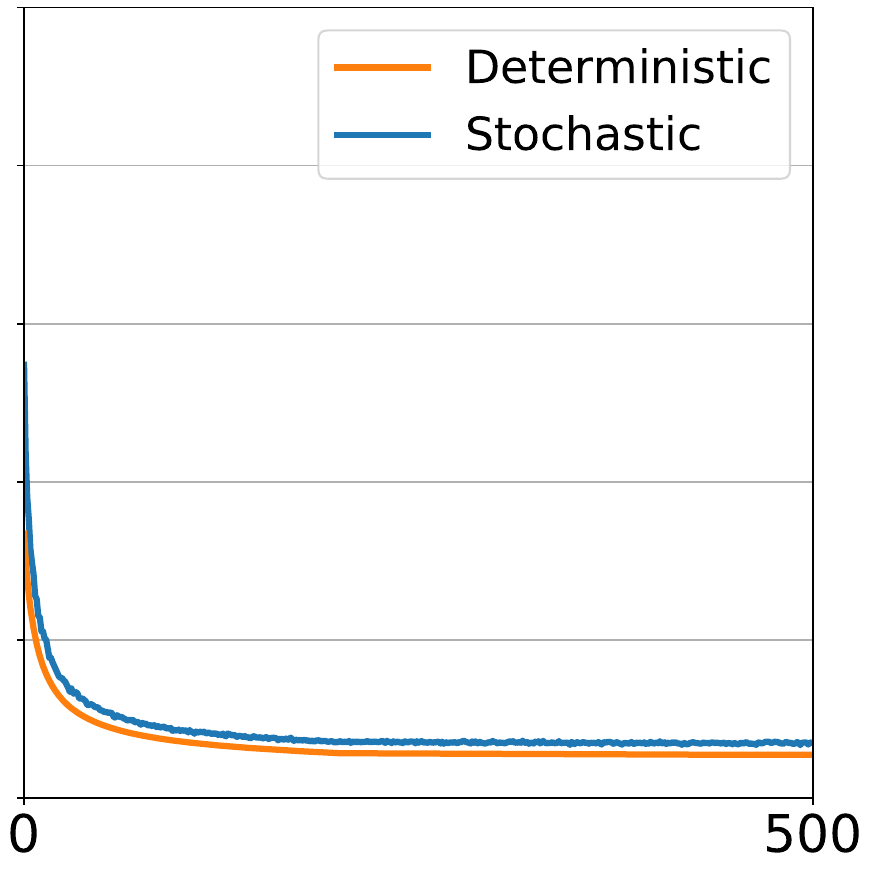}
        \caption{\textbf{\(\theta = 0.9\)}}
        \label{fig:synthetic-3-3}
    \end{subfigure}
    \begin{subfigure}[t]{0.18\textwidth}
        \centering
        \includegraphics[width=1\linewidth]{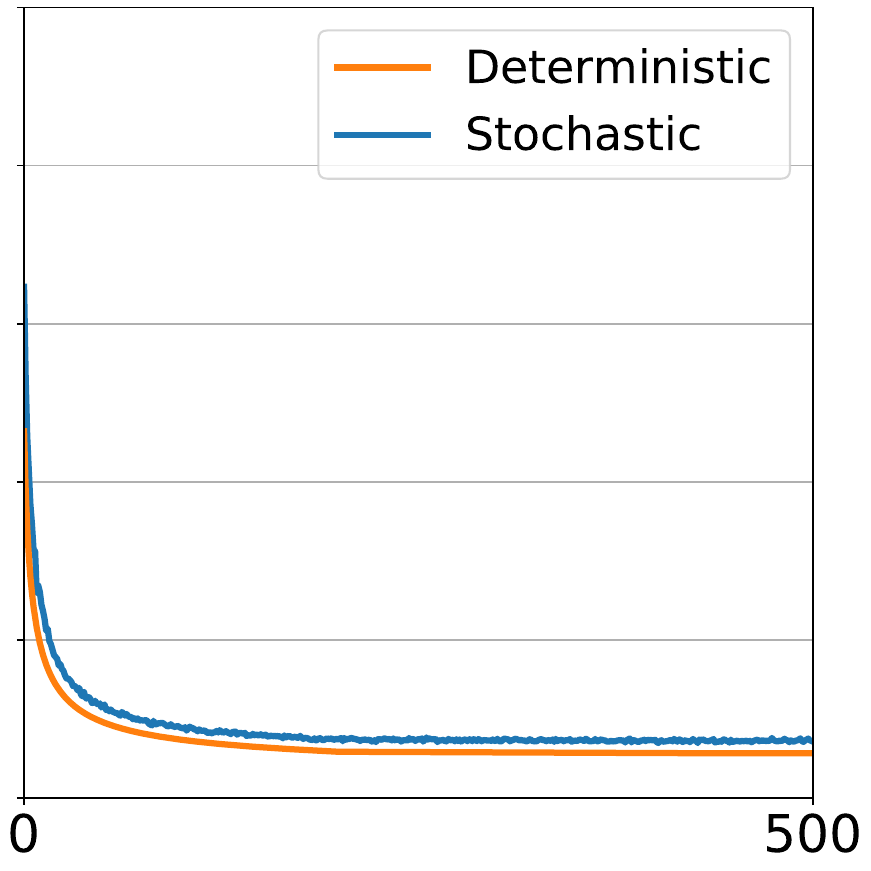}
        \caption{\textbf{\(\theta = 0.8\)}}
        \label{fig:synthetic-3-6}
    \end{subfigure}
    \begin{subfigure}[t]{0.18\textwidth}
        \centering
        \includegraphics[width=1\linewidth]{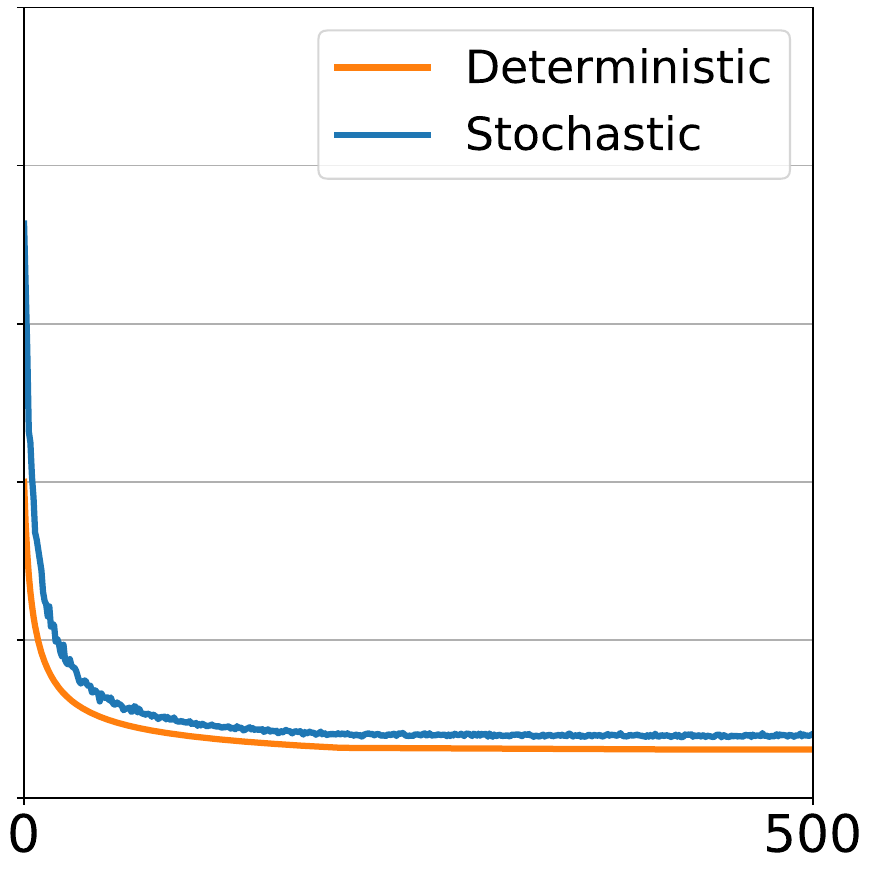}
        \caption{\textbf{\(\theta = 0.7\)}}
        \label{fig:synthetic-3-9}
    \end{subfigure}
    \begin{subfigure}[t]{0.18\textwidth}
        \centering
        \includegraphics[width=1\linewidth]{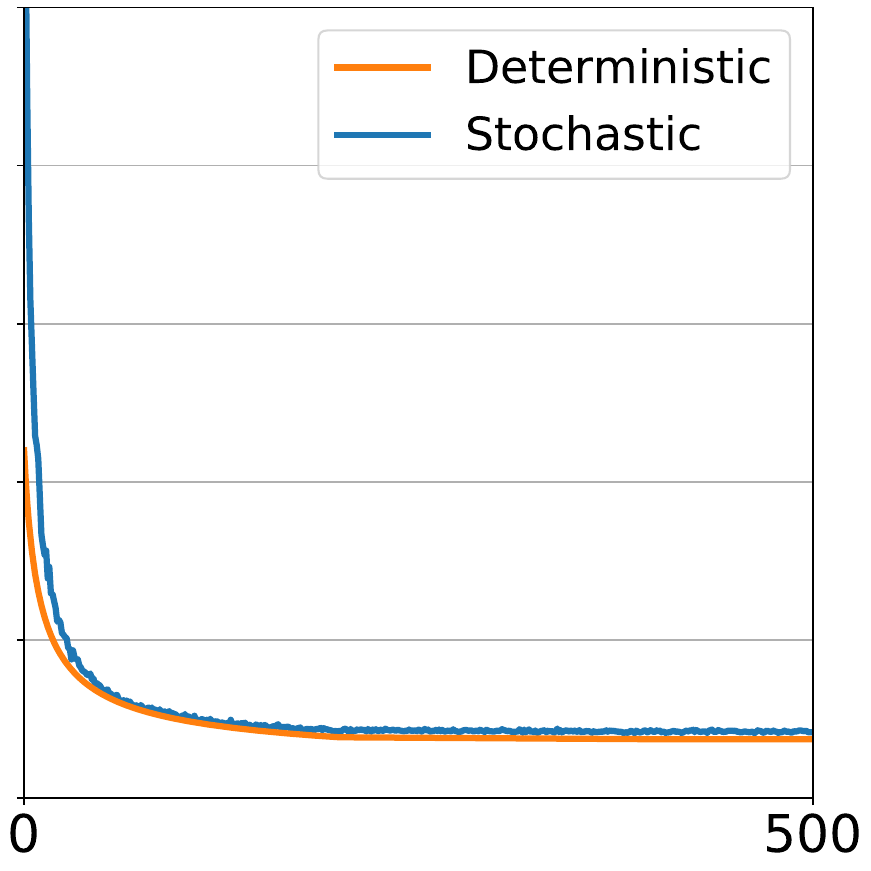}
        \caption{\textbf{\(\theta = 0.6\)}}
        \label{fig:synthetic-3-9}
    \end{subfigure}

    \caption{{\small\textbf{Experiment on synthetic data:} loss of the Tucker R-TRL as a function of the number of epochs for the stochastic version (orange) and the deterministic one based on the regularized objective function (blue). As expected, both formulations are empirically the same.}}
    \label{fig:synthetic_tucker}
\end{figure*}

Here, we empirically demonstrate the equivalence between our stochastic rank regularization and the deterministic regularization based formulation of the dropout.

To do so, we first created a random regression weight tensor \( \mytensor{W} \) to be a third order tensor of size \((25 \times 25 \times 25)\), formed as a rank--$10$ Kruskal tensor, the factors of which were sampled from an i.i.d. Gaussian distribution.
We then generated a tensor of \(5 000\) random samples, \mytensor{X} of size \((5 000 \times 25 \times 25 \times 25)\), the elements of which were sampled from a Normal distribution. Finally, we constructed the corresponding response array \myvector{y} of size \(5 000\) as: \( \forall i \in \myrange{1}{5 000}, \, \myvector{y_i} = \myinner{\mytensor{X}_i}{\mytensor{W}}\).
Using the same regression weight tensor and same procedure, we also generated \(1000\) testing samples and labels.

We use this data to train a CP R-TRL with our Tensor Dropout with rank \(15\), using both our Bernoulli stochastic formulation (equation 9 in main paper) and its deterministic counter-part (equation 10 in main paper). We train for \(500\) epochs, with a mini-batch size of \(100\), and an initial learning rate of \(10e-4\), which we decrease by a factor of \(10\) every \(200\) epochs. Figure~\ref{fig:synthetic} shows the loss function as a function of the epoch number. As expected, both formulations are identical.

\subsection{Large Scale Image Classification}
In this section, we report results for large-scale image classification on the ImageNet dataset.

\paragraph{ImageNet~\cite{imagenet}} is a large dataset for image classification composed of \(1.2\) million training images and \(50,000\) images for validation.
We evaluate the classification error in terms of top-1 and top-5 classification accuracy on a \(224 \times 224\) single center crop from the raw input images.
For training, we use a ResNet-101 and follow the same procedure and setting as \cite{he2016deep,kossaifi2018tensor}. Note that the original ResNet paper \cite{he2016deep} reports 10-crop validation error on ImageNet. Since then, single-crop has become the more common method of processing. The ResNet authors reported their single-crop results in their up-to-date code repository \cite{kaiminghe_2016}.

\begin{table}[!bht]
\caption{Classification accuracy on ImageNet with Bernoulli Tucker R-TRL. TRL ($\theta$ = 1) is the corresponding deterministic architecture.}
\begin{center}
\resizebox{0.8 \columnwidth}{!}{
\begin{tabular}{lllll}
\toprule
  & & \multicolumn{2}{c}{\bf Accuracy} \\
\cmidrule{3-4}
\multirow{-2}{*}{\textbf{Model}} & \multirow{-2}{*}{$\theta$} & \textbf{Top-1} (\%) & \textbf{Top-5} (\%) \\
\midrule
ResNet         &   \xmark  &  77.1  & 93.4  \\
TRL \cite{kossaifi2018tensor} &   1.0     &  77.1    & 93.5  \\
\textbf{Ours}           &   0.9     &  77.7    & 93.7  \\
\textbf{Ours}           &   0.8     &  77.7    & 93.7  \\
\textbf{Ours}           &   0.7     &  77.4    & 93.6  \\
\textbf{Ours}           &   0.6     &  \textbf{78.0}    & \textbf{93.8}  \\
\bottomrule
\end{tabular}
}
\end{center}
\label{table:imagenet}
\end{table}

The results, Table~\ref{table:imagenet}, show that our method outperforms the baselines, with higher classification accuracy even for large values of \(\theta\). This suggests the stochastic rank regularization helps constrain the learned representations to be more general. That leads to reduced over-fitting and better out-of-sample generalizability. The improvement of tensor dropout is greater than that of regular dropout, which has been reported to be around 0.3 percentage points in previous studies on ResNets trained on ImageNet \cite{ghiasi2018dropblock, kornblith2019better}.

\paragraph{Robustness study}
In addition to improving performance by reducing over-fitting, we demonstrate that our proposed stochastic regularization makes the model more robust to perturbations in the input, for both random noise and adversarial attacks.

In particular, we tested robustness to adversarial examples produced using three different types of attack: Fast Gradient Sign Method (FGSM) \cite{kurakin2016adversarial}, Basic Iterative Method (BIM)~\cite{kurakin2016adversarial} and Projected Gradient Descend (PGD)~\cite{madry2017towards} in Foolbox~\cite{rauber2017foolbox}. All attacks were untargeted. We note, that in~\cite{madry2017towards}, the authors showed that a method robust to PGD attack is reasonably robust to any form of first order gradient based attacks. In this method, the sign of the optimization gradient multiplied by the perturbation magnitude is added to the image in a single iteration. The perturbations we used are of magnitudes \(\lambda \times 10^{-3}, \lambda \in 
\{2, 8, 16\} \)
. For an input image normalized in the range [0, 1], $\lambda$ will be divided by 255. For the iterative methods BIM and PGD, we follow~\cite{kurakin2016adversarial} setting the step size to $1$ and the number of iterations to ${\text{min}(\lambda+4,1.25\lambda)}$. This fixed computational budget provides a compromise between the resources available to the attacker and the computational cost of the experiments.

As the results from Table~\ref{tab:imagenet-attacks} show, the proposed approach is significantly more robust to adversarial attacks across the entire range of adversarial noise tested. This further highlights that R-TRL is able to generalize better.

\begin{table}[!bht]
    \centering
    \small
    \caption{\textbf{Real-valued network performance on ImageNet} for FGSM, BIM and PGD attacks with~$\lambda \in \{2,8,16\}$. 
    We report classification accuracy in all cases.}
    \label{tab:imagenet-attacks}
    \resizebox{1 \columnwidth}{!}{
    \begin{tabular}{cccccc}
    \toprule[1.2pt]
    \multicolumn{2}{c}{\textbf{Attack}}  & \multicolumn{4}{c}{\textbf{Method}}       \\
    \cmidrule{3-6}
    \multirow{2}{*}{\textbf{Type}} & \multirow{2}{*}{\(\lambda\)}      &  \multirow{2}{*}{\textbf{Baseline}} & \multicolumn{3}{c}{\textbf{Ours}}\\
    \cmidrule{4-6}
    \addlinespace[0.1cm]
    &  &  &  \(\mathbf{\theta=0.8}\) &  \(\mathbf{\theta=0.7}\) &  \(\mathbf{\theta=0.6}\) \\
    \toprule[1pt]
    \multicolumn{2}{c}{\textbf{Clean}} & \multirow{2}{*}{$77.1$} & \multirow{2}{*}{$77.7$} & \multirow{2}{*}{ $77.4$ } & \multirow{2}{*}{ $\mathbf{78.0}$} \\
    \multicolumn{2}{c}{\emph{(no attack)}} &  &  & &  \\
    \midrule[0.8pt]
    \parbox[t]{2mm}{\multirow{3}{*}{\rotatebox[origin=c]{90}{\textbf{FGSM}}}}
        & \textbf{2 } &  $26.1$ & $26.0$ & $35.3$ & $\mathbf{47.4}$  \\
        & \textbf{8 } &  $11.1$ & $15.4$ & $26.0$ & $\mathbf{41.6}$  \\
        & \textbf{16} &  $8.5$ & $14.1$ & $24.0$ & $\mathbf{39.5}$  \\
    \midrule[0.8pt]
    \parbox[t]{2mm}{\multirow{3}{*}{\rotatebox[origin=c]{90}{\textbf{BIM}}}}
        & \textbf{2 } &  $26.1$ & $26.0$ & $35.3$ & $\mathbf{47.3}$  \\
        & \textbf{8 } &  $1.0$ & $4.0$ & $9.9$ & $\mathbf{26.1}$  \\
        & \textbf{16} &  $0.1$ & $1.0$ & $2.8$ & $\mathbf{13.2}$  \\
        \midrule[0.8pt]
    \parbox[t]{2mm}{\multirow{3}{*}{\rotatebox[origin=c]{90}{\textbf{PGD}}}}
        & \textbf{2 } &  $26.0$ & $26.0$ & $35.6$ & $\mathbf{47.1}$  \\
        & \textbf{8 } &  $0.8$ & $4.5$ & $11.3$ & $\mathbf{27.9}$  \\
        & \textbf{16} &  $0$ & $1.4$ & $3.8$ & $\mathbf{13.5}$  \\
    \bottomrule[1.2pt]
    \end{tabular}
    }
\end{table}

\begin{figure}[!htb]

    \centering
        \subcaptionbox*{}{
        \raisebox{90pt}{
        \rotatebox[origin=t]{90}{Classification accuracy (\%)}
        }
    }
    \includegraphics[width=0.85\linewidth]{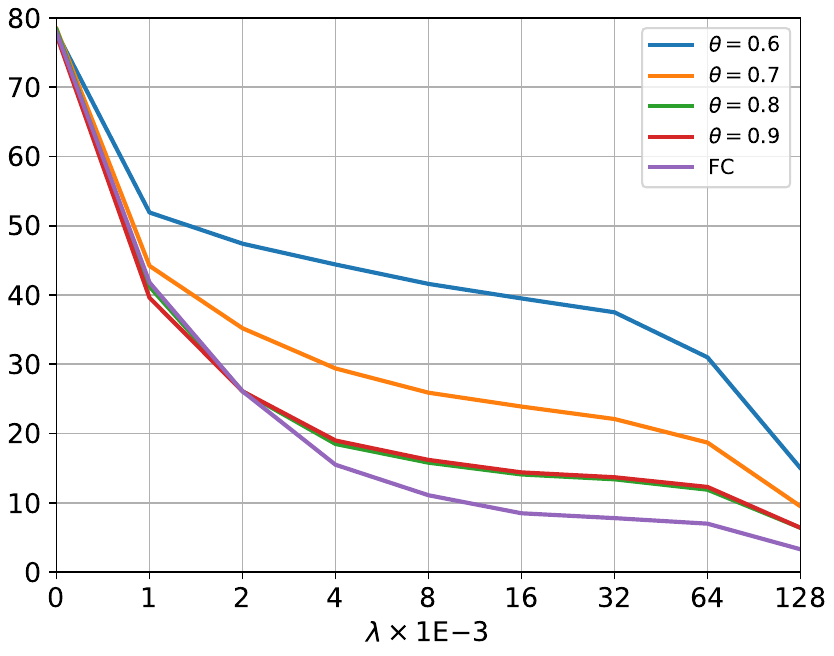}
    \vspace{-15pt}
    \caption{{\small\textbf{Robustness to adversarial attacks on ImageNet}}. Our R-TRL architecture is much more robust to adversarial attacks, even though adversarial training was not used.}
    \label{fig:imagenet-attacks}
\end{figure}

Moreover, in figure~\ref{fig:imagenet-attacks}, we report the classification accuracy as a function of the added adversarial noise. Specifically, following~\cite{brendel2017decision}, we sampled \(1,000\) unseen test images. The models were trained \emph{without} any adversarial training and adversarial noise was added using the Fast Gradient Sign method. Our Tucker R-TRL model is much more robust to adversarial attacks.

Adversarial defenses through randomness have been explored before, in different forms. For example, perturbing the input with isotropic Gaussian noise gives substantial and certifiable robustness to attacks on ImageNet \cite{pmlr-v97-cohen19c}, and adding random noise to each layer of the network provides defense against strong attacks \cite{liu2018towards}. However, these prior methods differ fundamentally to the one we present here. The randomness in our method is concerned with the rank of the factor tensors during training only and not perturbing the input or weights. Theoretically, our method is more related to regular dropout as we show in section \ref{sec:cifar100}.

\subsection{Phenotypic trait prediction from MRI data}
Contemporary neurological research often uses scalar measures of brain or grey matter volumes over time as a proxy for brain neuronal volume. For example, to provide estimates of neuronal loss over time \cite{lemaitre2012normal}. However, the global and local structure of brain tissue also change \cite{kovalev2001three}, causing structural and textural changes that are not captured by scalar volume measures. CNNs are more suited for learning these intricate and complex relationships. However, the highly multi-dimensional mappings come with added risk of overfitting. Model generalizability is therefore paramount in this domain, providing a concrete demonstration of the strengths of the tensor dropout.

In a regression setting, we investigate the performance of our method on a very large MRI dataset on the task of age prediction.

The difference between an individual's chronologial age and the age as predicted by a trained model is often referred to as \emph{brain-age}. This metric which has been associated with diseases \cite{franke2012brain} and increased risk of mortality \cite{cole2017brain}. A more accurate and robust metric of brain condition can consequently lead to more accurate disease diagnoses.

In addition to the general neuroradiological motivations outlined above, this task is particularly interesting since MRI volumes are large 3D tensors, all modes of which carry important information. The spatial information is traditionally discarded during the flattening process, which we avoid by using a tensor regression layer. In these experiments we train the entire model, but any pre-trained network can be easily modified post-hoc to make use of the TRL.

\paragraph{UK Biobank brain MRI dataset \cite{sudlow2015uk}} is the world's largest MRI imaging database of its kind. The aim of the UK Biobank Imaging Study is to capture MRI scans of vital organs for \(100,000\) primarily healthy individuals by 2022. Associations between these images and lifestyle factors and health outcomes, both of which are already available in the UK Biobank, will enable researchers to improve diagnoses and treatments for numerous diseases.

The data we use here consists of T1-weighted \(182\times218\times182\) MR images of the brain captured on a 3 T Siemens Skyra system. \(11,500\) are used for training, \(3,800\) are used for validation and \(3,800\) samples are used to test. The target label is the age for each individual at the time of MRI capture. We use skull-stripped images that have been aligned to the MNI152 template \cite{jenkinson2002improved} for head-size normalization. We then center and scale each image to zero mean and unit variance for intensity normalization.

\begin{table}[hbt]
\centering
\caption{\textbf{Classification accuracy for UK Biobank MRI.} The ResNet models with R-TRL significantly outperforms the version with a fully-connected (FC) layer.}
\label{table:mri}
\resizebox{0.9 \columnwidth}{!}{
\begin{tabular}{lll}
\toprule
\multicolumn{1}{c}{\bf Architecture} &  \multicolumn{1}{c}{\textbf{Regression}} & \multicolumn{1}{c}{\bf MAE}\\
\midrule
3D-ResNet    &  FC  & 2.96 years \\ %
3D-ResNet &  Tucker  & 2.70 years \\ %
\textbf{Ours} &  Randomized Tucker  & \textbf{2.65 years} \\ %
\textbf{Ours} & Randomized CP & \textbf{2.58 years} \\
\bottomrule
\end{tabular}
}

\end{table}

\paragraph{Results:} For MRI-based experiments we implement an 18-layer ResNet with three-dimensional convolutions. We minimize the mean squared error using Adam \cite{kingma2014adam}, starting with an initial learning rate of \(10^{-4}\), reduced by a factor of 10 at epochs 25, 50, and 75. We train for 100 epochs with a mini-batch size of 8 and a weight decay (\(\mathrm{L}_{2}\) penalty) of \(5\times10^{-4}\). For Tucker-based R-TRL we used a tensor with rank \(128 \times 6 \times 7 \times 6\). For CP-based R-TRL we used a Kruskal tensor with \(82\) components.
 As previously observed, our randomized tensor regression network outperforms the 3D-ResNet baseline by a large margin, Table~\ref{table:mri}.
To put this into context, the current state-of-art for convolutional neural networks on age prediction from brain MRI on most datasets is an MAE of around 3.6 years \cite{cole2017predicting}.
\paragraph{Robustness study:}
We tested the robustness of our model to  white Gaussian noise added to the MRI data. Noise in MRI data typically follows a Rician distribution but can be approximated by a Gaussian for signal-to-noise ratios (SNR) greater than \(2\) \cite{gudbjartsson1995rician}. As both the signal (MRI voxel intensities) and noise are zero-mean, we define \( \mathrm{SNR} = \frac{\sigma_{\mathrm{signal}}^2}{\sigma_{\mathrm{noise}}^2} \), where \( \sigma \) is the variance. We incrementally increase the added noise in the test set and compare the error rate of the models.

\begin{figure}[!hbt]
    \centering
    \includegraphics[width=0.9\linewidth]{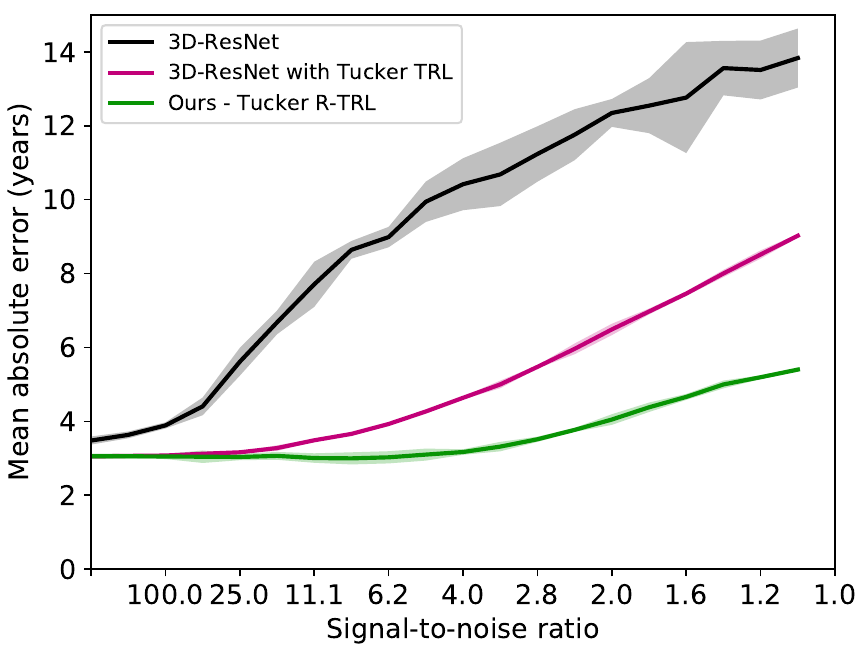}
    \caption{\textbf{Age prediction error on the MRI test set} as a function of increased added Gaussian noise. Shaded regions indicate \(95\%\) confidence intervals for \(5\) independent runs. }
    \label{fig:mri-noise}
\end{figure}

\begin{figure*}[!htbp]
    \centering
        \subcaptionbox*{}{
        \raisebox{60pt}{
            \rotatebox[origin=t]{90}{\small{Classification accuracy (\%)}}
        }
    }
    \begin{subfigure}[t]{0.29\textwidth}
        \hspace{-10pt} \includegraphics[width=1.05\linewidth]{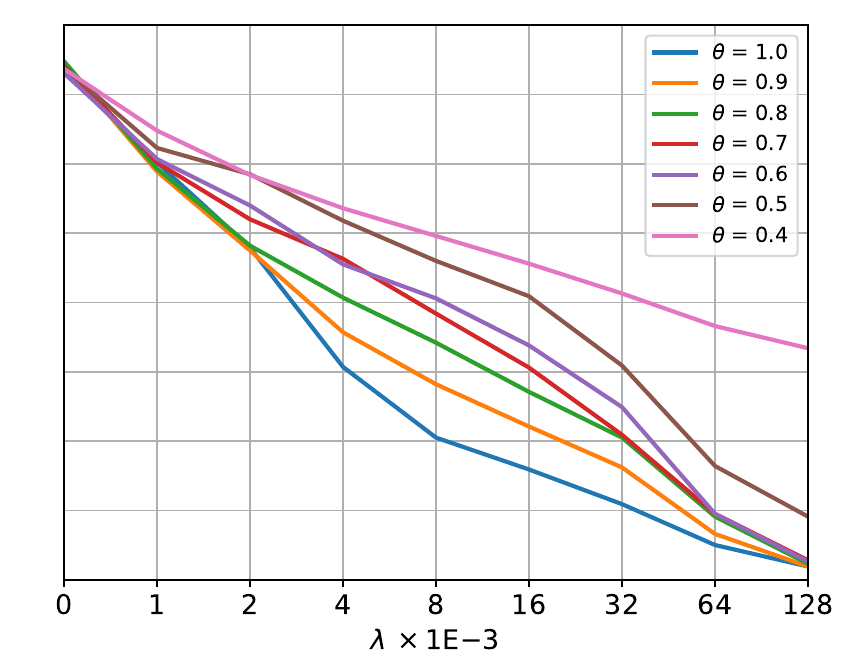}
        \caption{FGS attack on Bernoulli Tucker R-TRL with different drop rates.}
        \label{fig:bernoulli-adversarial-tucker}
    \end{subfigure}
    \hfill
    \begin{subfigure}[t]{0.29\textwidth}
        \hspace{-10pt} \includegraphics[width=1.05\linewidth]{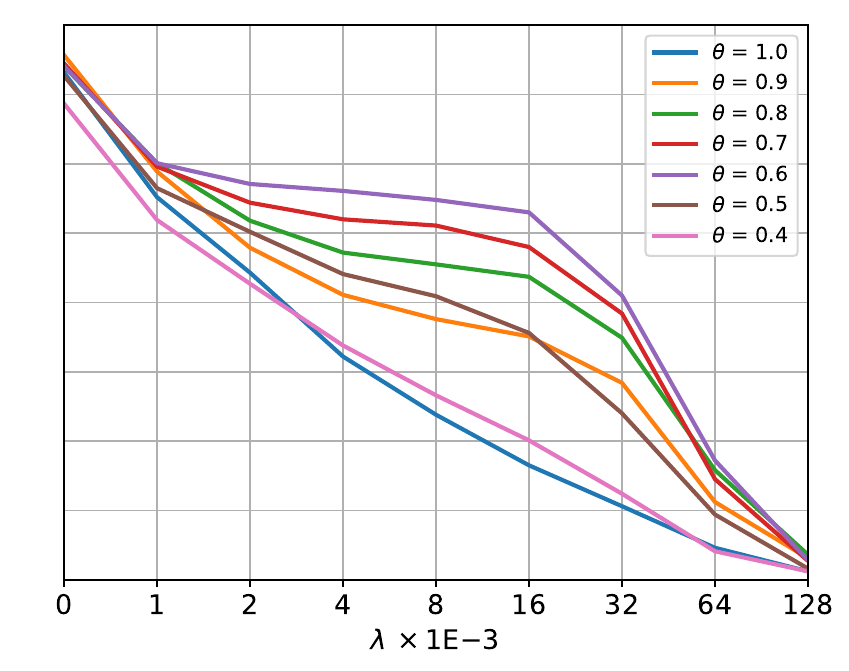}
        \caption{FGS attack on CP R-TRL with different drop rates.}
        \label{fig:bernoulli-adversarial-cp}
    \end{subfigure}
    \hfill
    \begin{subfigure}[t]{0.29\textwidth}
        \hspace{-10pt} \includegraphics[width=1.05\linewidth]{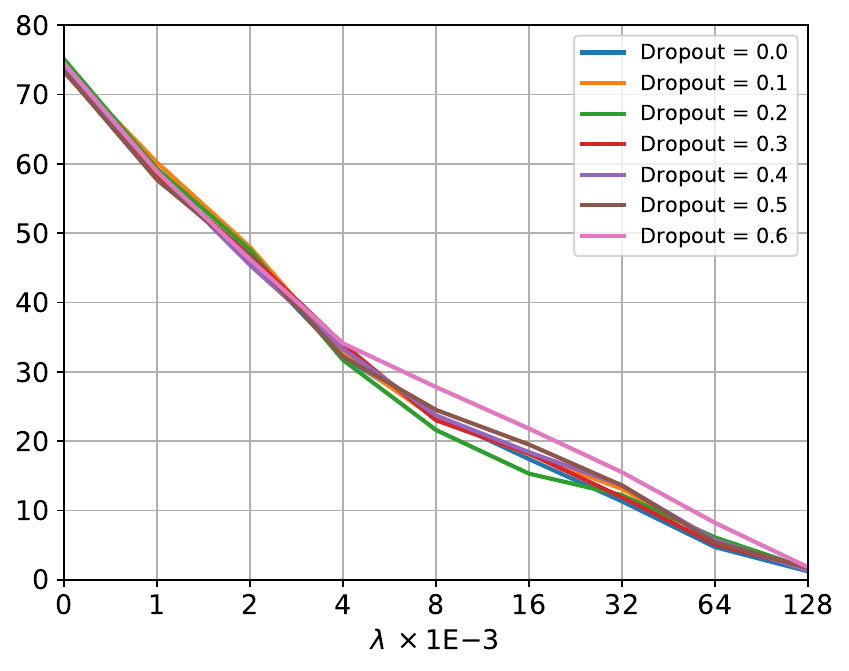}
        \caption{FGS attack on Tucker TRL with different dropout rates.}
        \label{fig:adversarial_dropout_attack}
    \end{subfigure}
    \caption{\textbf{Robustness to adversarial attacks}, measured by adding adversarial noise to the test images, using the Fast Gradient Sign, on CIFAR-100 and Bernoulli drop. We compare a Tucker tensor regression layer with dropout applied to the regression weight tensor (Subfig.~\ref{fig:adversarial_dropout_attack}) to our randomized TRL, both in the Tucker (Subfig.~\ref{fig:bernoulli-adversarial-tucker}) and CP (Subfig.~\ref{fig:bernoulli-adversarial-cp}) case. Our approach is more robust.}
    \label{fig:adversarial_attack}
\end{figure*}

The ResNet with R-TRL is significantly more robust to added white Gaussian noise compared to the same architectures without it (figure~\ref{fig:mri-noise}). At signal-to-noise ratios below 10, the accuracy of a standard fully-connected ResNet is worse than a naive model that predicts the mean of training set (MAE = \(7.9\) years).
Brain morphology is an important attribute that has been associated with various biological traits including cognitive function and overall health \cite{pfefferbaum1994quantitative,swan1998association}. By keeping the structure of the brain represented in MRI in every layer of the architecture, the model has more information to learn a more accurate representation of the entire input.
Randomly dropping the rank forces the representation to be robust to confounds. This a particularly important property for MRI analysis since intensities and noise artifacts can vary significantly between MRI scanners \cite{wang1998correction}. Randomized tensor regression layers enable both more accurate and more robust trait predictions from MRI that can consequently lead to more accurate disease diagnoses.

\subsection{Ablation Studies on CIFAR-100}
\label{sec:cifar100}
In the image classification setting, we perform a thorough study of our method on the CIFAR-100 dataset. We empirically compare our approach to both standard baseline, traditional tensor regression, and regular dropout, and assess the robustness of each method in the face of adversarial noise.

\paragraph{CIFAR-100~\cite{krizhevsky2009learning}} consists of 60,000 \(32 \times 32\) RGB images in 100 classes, divided into \(50,000\) images for training and \(10,000\) for testing. We pre-processed the data by centering and scaling each image and then augmented the training images with random cropping and random horizontal flipping.

We compare the randomized tensor regression layer to full-rank tensor regression, average pooling and a fully-connected layer in an 18-layer residual network (ResNet) \cite{he2016deep}. For all networks, we used a batch size of \(128\) and trained for \(400\) epochs, and minimized the cross-entropy loss using stochastic gradient descent (SGD). The initial learning rate was set to \(0.01\) and lowered by a factor of \(10\) at epochs \(150\), \(250\) and \(350\). We used a weight decay (\(\mathrm{L}_{2}\) penalty) of \(10^{-4}\) and a momentum of \(0.9\).

\paragraph{Classification results:}
Table~\ref{table:cifar100} presents results obtained on the CIFAR-100 dataset, on which our method matches or outperforms other methods, including the same architectures without R-TRL.
Our regularization method makes the network more robust by reducing over-fitting, thus allowing for superior performance on the testing set.

\begin{table}[t]
\caption{\textbf{Classification accuracy for CIFAR-100} with a ResNet and various regression layers for classification.}
\label{table:cifar100}
\centering
\resizebox{0.7 \columnwidth}{!}{
\begin{tabular}{ll}
\toprule
\multicolumn{1}{c}{\bf ResNet classification}  & \multicolumn{1}{c}{\bf Accuracy}\\ %
\midrule
FC     &75.88  \%\\ %
FC + dropout  &75.84  \%\\ %
Tucker  &76.02 \%\\ %
CP  &75.77 \%\\
\textbf{Randomized Tucker}   &76.05 \%\\ %
\textbf{Randomized CP}   & \textbf{76.19} \%\\ %
\bottomrule
\end{tabular}
}
\end{table}

\begin{figure*}[!htbp]
    \centering
    \quad
    \begin{subfigure}[t]{0.45\textwidth}
        \centering
        \includegraphics[width=1\linewidth]{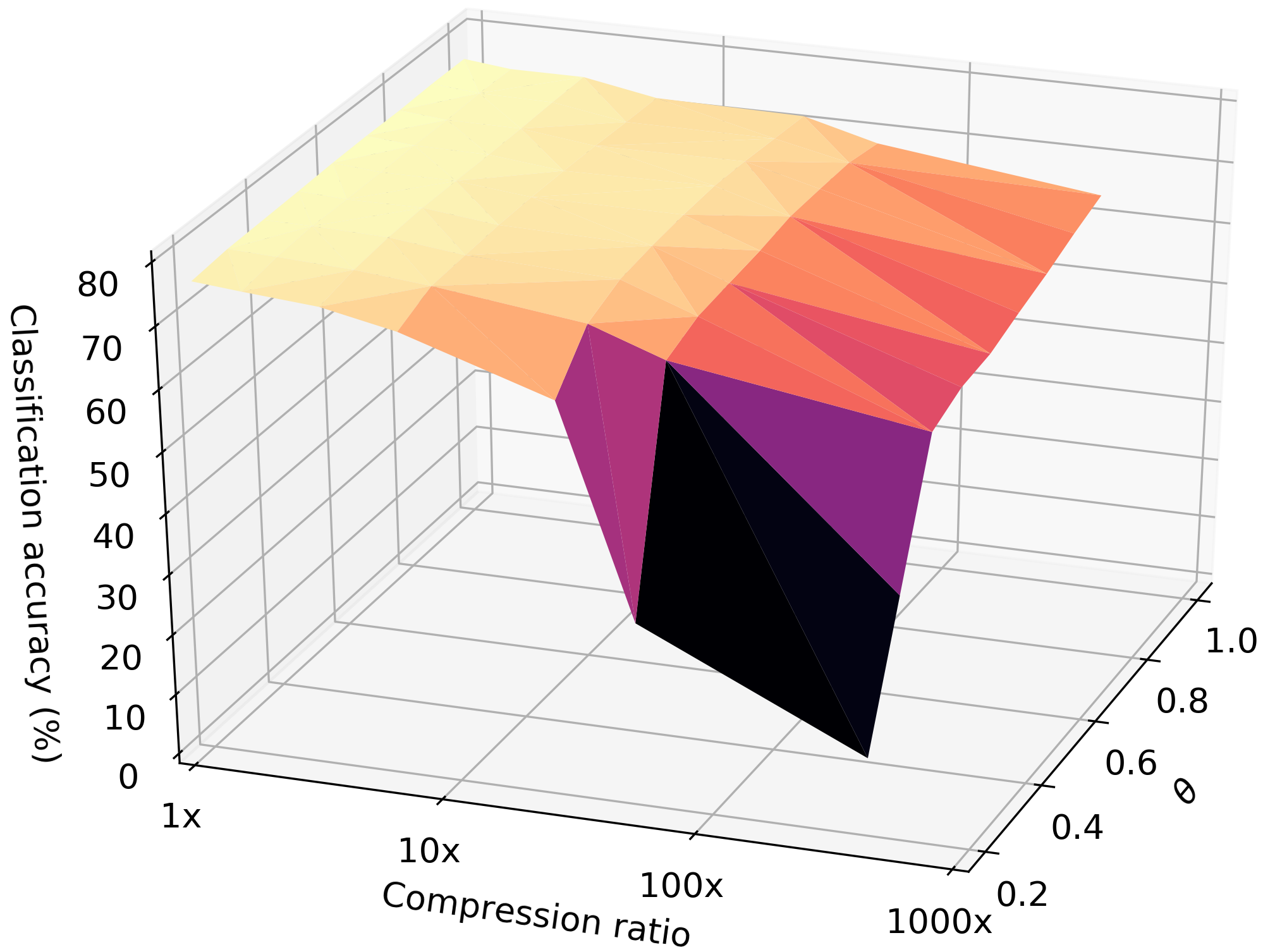}
        \caption{Bernoulli R-TRL}
        \label{fig:bernoulli-surf}
    \end{subfigure}
    \hfill
    \begin{subfigure}[t]{0.45\textwidth}
        \centering
        \includegraphics[width=1\linewidth]{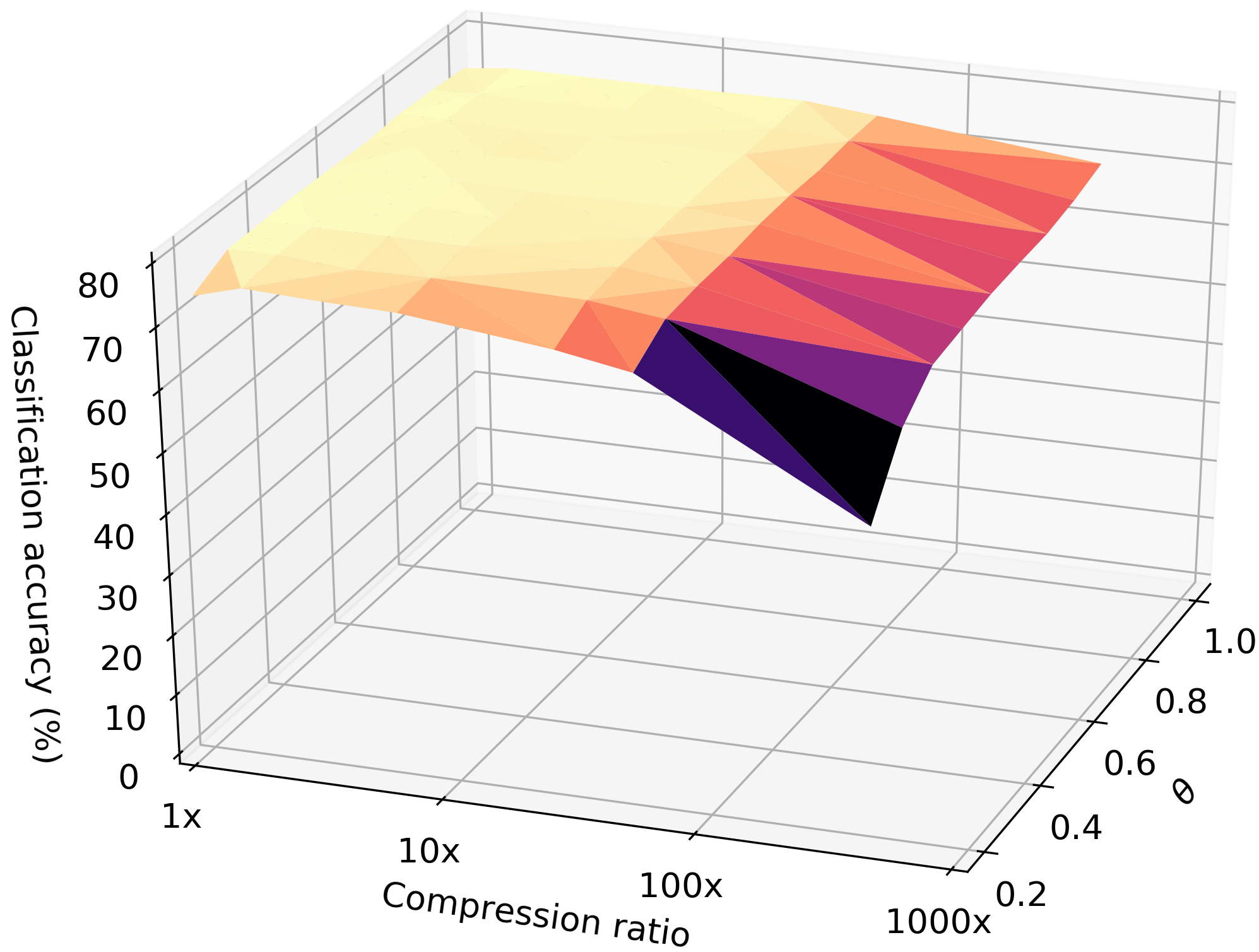}
        \caption{R-TRL with replacement}
        \label{fig:replacement-surf}
    \end{subfigure}
    \quad
    \caption{\textbf{CIFAR-100 test accuracy}, as a function of the compression ratio (logarithmic scale) and drop rate $\theta$. There is a large region for which reducing both the rank and $\theta$ does not hurt performance. 
    }
    \label{fig:surface}
\end{figure*}

A natural question is whether the model is sensitive to the choice of rank and $\theta$ (or drop rate when sampling with repetition). To assess this, we show the performance as a function of both rank and $\theta$ in figure~\ref{fig:bernoulli-surf}. The reduction in rank is presented as the \(\text{compression ratio} = \frac{\text{size of full tensor}}{\text{size of factorized cores and factors}}\). As can be observed, there is a large surface for which performance remains the same while decreasing both parameters (note the logarithmic scale for the rank). This means that, as demonstrated, choosing good values for CIFAR-100 was not a problem. Rank insensitivity cannot be guaranteed across all tasks and datasets, but rank-selection was not found to be an issue for ImageNet or MRI tasks.

\paragraph{Performance as a function of rank and $\theta$ in replacement R-TRL}\label{sec:perforanace-replacement-rtrl}

To illustrate the generality of our approach, which does not depend on the Bernoulli sampling, we perform a similar experiment with a different randomization: instead of using a Bernoulli random variable, we sample components with replacement according to a uniform sampling matrix (figure ~\ref{fig:replacement-surf}). As for the Bernoulli case, there is a large surface for which performance remains the same while decreasing both parameters.

\paragraph{Comparison with regular dropout}
One question is whether our proposed tensor dropout is more robust than traditional dropout applied directly to the weights. To test this, we apply FGSM adversarial perturbations to each method, with varying magnitudes \(\lambda \times 10^{-3}, \lambda \in \{1, 2, 4, 8, 16, 32, 64, 128\} \). We sample \(1,000\) images from the test set \cite{brendel2017decision}. The models were trained \emph{without} any adversarial training, on the training set, and adversarial noise was added to the test samples using the Fast Gradient Sign method. The results of which can be seen in figure~\ref{fig:adversarial_attack}. Our model is much more robust to adversarial attacks. Intuitively our method is able to leverage redundancies in the latent subspace, without creating holes in the weights, unlike dropout. In addition, since the randomization is used only during training (and not at test time), this forces the latent decomposition to be over-complete and account for noise, thus rendering the model more robust to perturbation.

\section{Conclusion}
We introduced tensor dropout, a novel randomized tensor decomposition, suitable for end-to-end training of tensor regression layers. Adding stochasticity on the rank during training renders the network significantly more robust and leads to better performance. This results in networks that are more resilient to noise, both adversarial and random, without any addition such as adversarial training. Our results demonstrate superior performance on a variety of real-life, large-scale challenging tasks, including MRI data and images, as well as increased robustness. Finally, we establish the link between this randomized TRL and dropout on the deterministic TRL.

\appendices

\section*{Acknowledgment}
This research has been conducted using the UK Biobank Resource under Application
Number 18545.

\ifCLASSOPTIONcaptionsoff
  \newpage
\fi

\bibliographystyle{ieeetr}
\bibliography{refs}

\newpage
\onecolumn

\section{Connection between stochastic and deterministic regression}\label{sec:minimisation-proof}

In this section, we study the empirical risk minimization of Tensor Dropout. We establish the link between Tensor Dropout with stochastic Tucker decomposition and the deterministic regularized loss. We do this by proving the equality between equations \ref{eq:rnd-trl} and \ref{eq:rnd-trl-tucker} (Theorem~\ref{thm:tucker}).

\begin{proof}[Proof of Theorem~\ref{thm:tucker}]
The minimization objective in equation~\ref{eq:rnd-trl} can be reformulated by expanding the tensor contractions in Tucker form as:
\begin{equation}\label{eq:tucker-exp}
\begin{split}
\mytensor{\tilde W}_{i_{0}, \dots,i_{N}}
    =
    \sum_{r_{0}=1}^{R_{0}}
    \cdots
    \sum_{r_{N}=1}^{R_{N}} 
    \myvector{g}_{r_{0}, \dots,r_{N}}
    \lambda_{r_0}^{(0)} \dots \lambda_{r_{N}}^{(N)} 
    \mymatrix{U}_{i_{0}r_{0}}^{(0)} \dots \mymatrix{U}_{i_{N}r_{N}}^{(N)}
\end{split}
\end{equation}

By considering the expectation over the empirical risk, we get:
\begin{equation*}
\begin{split}
\myexpectation{\myvector{\lambda}}{\frac{1}{S-1}\sum^{S-1}_{k=0}
\left(\myvector{y}^{(k)} - \myinner{\mytensor{\tilde W}}{\mytensor{X}^{(k)}}
\right)^2}
&=
\frac{1}{S-1}\sum^{S-1}_{k=0}
\myexpectation{\myvector{\lambda}}{\left(\myvector{y}^{(k)} - \myinner{\mytensor{\tilde W}}{\mytensor{X}^{(k)}}
\right)^2} \\
&=
\underbrace{
\frac{1}{S-1}\sum^{S-1}_{k=0}
\myexpectation{\myvector{\lambda}}{\left(
\myvector{y}^{(k)} - \myinner{\mytensor{\tilde W}}{\mytensor{X}^{(k)}}
\right)}^2
}_\textrm{E} 
+ \underbrace{
\frac{1}{S-1}\sum^{S-1}_{k=0}
\myvar{\myvector{y}^{(k)} - \myinner{\mytensor{\tilde W}}{\mytensor{X}^{(k)}}}
}_\textrm{V}
\end{split}
\end{equation*}

The expectation E and variance V can then be written separately, using the Tucker expansion from Equation~\ref{eq:tucker-exp}:
\begin{equation*}
\begin{split}
\textrm{E} & =
\frac{1}{S-1}
\sum^{S-1}_{k=0}{\left(
\myvector{y}^{(k)} - 
\sum_{i_{0}}\cdots\sum_{i_{N}}
\mytensor{X}_{i_{0}, \dots,i_{N}}^{(k)}{
\sum_{r_{0}}
\cdots
\sum_{r_{N}}
\myvector{g}_{r_{0}, \dots,r_{N}}
\myexpectation{}{    \lambda_{r_0}^{(0)} \dots \lambda_{r_{N}}^{(N)}}
}
\mymatrix{U}_{i_{0}r_{0}}^{(0)} \dots \mymatrix{U}_{i_{N}r_{N}}^{(N)}
\right)
}^2 \\
& = \frac{1}{S-1}\sum^{S-1}_{k=0}
\left(\myvector{y}^{(k)} - \theta^{N}\myinner{\mytensor{W}}{\mytensor{X}^{(k)}}
\right)^2
\text{ as } \lambda^{(0)}, \dots,  \lambda^{(N)} \text{ are i.i.d. }
\end{split}
\end{equation*}

\begin{equation*}
\begin{split}
\textrm{V} & =
\frac{1}{S-1}
\sum^{S-1}_{k=0}{\left(
{(-1)}^2
\sum_{i_{0}}\cdots\sum_{i_{N}}
{(\mytensor{X}_{i_{0}, \dots,i_{N}}^{(k)})}^2
\sum_{r_{0}}
\cdots
\sum_{r_{N}}
\myvector{g}_{r_{0}, \dots,r_{N}}^2
\myvar{\lambda_{r_0}^{(0)} \dots \lambda_{r_{N}}^{(N)}}
)
(\mymatrix{U}_{i_{0}r_{0}}^{(0)})^2 \dots (\mymatrix{U}_{i_{N}r_{N}}^{(N)})^2
\right)
} \\
& =
\frac{1}{S-1}
\theta^{N}(1-\theta^{N})
\sum^{S-1}_{k=0}\left(
{\sum_{i_{0}}\cdots\sum_{i_{N}}
(\mytensor{X}_{i_{0}, \dots,i_{N}}^{(k)})}^2
\mytensor{G}^{\star 2}
\times_0 (\mymatrix{U}^{(0)})^{\star 2} \dots
\times_{N} (\mymatrix{U}^{(N)})^{\star 2}
\right) \\
& =
\frac{\theta^{N}(1-\theta^{N})}{S-1}
\sum^{S-1}_{k=0}
\myinner{\mytensor{G}^{\star 2}
\times_0 (\mymatrix{U}^{(0)})^{\star 2} \dots
\times_{N} (\mymatrix{U}^{(N)})^{\star 2}}{(\mytensor{X}^{(k)})^{\star 2}})
\end{split}
\end{equation*}

Therefore, 
\begin{equation*}
\begin{split}
 \mathbb{E}_{ \, \myvector{\lambda}} \bm{ \big[ }\,
    \frac{1}{S-1}\sum^{S-1}_{k=0}
    \left(\myvector{y}^{(k)} - \myinner{\mytensor{\tilde W}}{\mytensor{X}^{(k)}}
    \right)^2
\,\bm{ \big] }
= \,\,
    & \frac{1}{S-1}\sum^{S-1}_{k=0}
    \left(\myvector{y}^{(k)} - \theta^{N}\myinner{\mytensor{W}}{\mytensor{X}^{(k)}}
    \right)^2 \\
    & + \frac{\theta^{N}(1-\theta^{N})}{S-1} \sum^{S-1}_{k=0}
    \myinner{\mytensor{G}^{\star 2}
    \times_0 (\mymatrix{U}^{(0)})^{\star 2} \dots
    \times_{N} (\mymatrix{U}^{(N)})^{\star 2}}{(\mytensor{X}^{(k)})^{\star 2}}.
\end{split}
\end{equation*}

\end{proof}

\section{}\label{sec:bernoulli-proof}

In this section, we study Tensor Dropout's empirical risk minimization and establish the link between Tensor Dropout with stochastic CP decomposition and the corresponding deterministic regularized loss (Theorem~\ref{thm:CPminimisation}). To do so, we first establish the equality between \ref{eq:bernouilli-rnd-weight-long} and \ref{eq:bernouilli-rnd-weight}. 
We then prove Theorem~\ref{thm:CPminimisation}, which is the equivalence of equations \ref{eq:rnd-trl} and \ref{eq:CPminimisation}.

\begin{proof}[CP decomposition with tensor dropout]
Here we prove the following equivalence (equations \ref{eq:bernouilli-rnd-weight-long} and \ref{eq:bernouilli-rnd-weight}):
\begin{equation}\label{eq:bernouilli-rnd-weight-proof}
\begin{split}
    \mytensor{\tilde W} & = 
    \sum_{k=0}^{R-1} \mymatrix{\tilde U}^{(0)}_k \circ \cdots \circ \mymatrix{\tilde U}^{(N)}_k\\
    & = 
    \sum_{k=0}^{R-1} \lambda_k \mymatrix{U}^{(0)}_k \circ \cdots \circ \mymatrix{U}^{(0)}_N \\
    & =
    \mykruskal{\myvector{\lambda};\, \mathbf{U}^{(0)}, \cdots, \mathbf{U}^{(N)}}
\end{split}
\end{equation}

This can be seen by looking at the individual elements of the sketched factors.
Let \(k \in \myrange{0}{N}\) and \(i_k \in \myrange{0}{I_k}, r \in \myrange{0}{R-1}\). Then 
\[
\mymatrix{\tilde U}_{i_k, r}^{(k)} = \sum_{j=0}^{R-1} \mymatrix{U}_{i_k, j}^{(k)}\mymatrix{M}_{j, r}.
\]
Since \(\mymatrix{M} = \text{diag}(\myvector{\lambda})\), \myie \(\forall i, j \in \myrange{0}{R-1}, \mymatrix{M}_{ij} = 0\) if \(i \neq j\), and \( \lambda_i\) otherwise, we get 
\[
\mymatrix{\tilde U}_{i_k, r}^{(k)} = \lambda_r \mymatrix{U}_{i_k, r}^{(k)}.
\]
It follows that \(\mytensor{\tilde W}_{i_0, i_1, \cdots, i_N} =
\sum_{r=0}^{R-1} \lambda_k \mymatrix{U}^{(0)}_k \circ \cdots \circ \lambda_k \mymatrix{U}^{(N)}_k
\)
Since \(\lambda_r \in \{0, 1\},\) we have 
\[
  \mytensor{\tilde W}_{i_0, i_1, \cdots, i_N} = 
  \sum_{r=0}^{R-1} \lambda_k \left(\mymatrix{U}^{(0)}_k \circ \cdots \circ \mymatrix{U}^{(N)}_k\right)
\]

\end{proof}

\begin{proof}[Proof of Theorem~\ref{thm:CPminimisation}]
Here we prove that Tensor Dropout with stochastic CP decomposition is equivalent to deterministic regularized loss.
By considering the expectation over the empirical risk, we get:
\begin{equation*}
\begin{split}
\myexpectation{\myvector{\lambda}}{\frac{1}{S-1}\sum^{S-1}_{k=0}
\left(\myvector{y}^{(k)} - \myinner{\mytensor{\tilde W}}{\mytensor{X}^{(k)}}
\right)^2}
&=
\frac{1}{S-1}\sum^{S-1}_{k=0}
\myexpectation{\myvector{\lambda}}{\left(\myvector{y}^{(k)} - \myinner{\mytensor{\tilde W}}{\mytensor{X}^{(k)}}
\right)^2} \\
&=
\underbrace{
\frac{1}{S-1}\sum^{S-1}_{k=0}
\myexpectation{\myvector{\lambda}}{\left(
\myvector{y}^{(k)} - \myinner{\mytensor{\tilde W}}{\mytensor{X}^{(k)}}
\right)}^2
}_\textrm{E} 
+ \underbrace{
\frac{1}{S-1}\sum^{S-1}_{k=0}
\myvar{\myvector{y}^{(k)} - \myinner{\mytensor{\tilde W}}{\mytensor{X}^{(k)}}}
}_\textrm{V}
\end{split}
\end{equation*}

For Tensor dropout where weights are constructed with CP factors: 
\( \mytensor{\tilde W} = \sum_{r=0}^{R-1}\lambda_{n} \mymatrix{U}_{i_{0}r_{0}}^{(0)} \dots \mymatrix{U}_{i_{N}r_{N}}^{(N)}\)
the minimization objective can be written as:

\begin{equation*}
\begin{split}
\textrm{E} & =
\frac{1}{S-1}
\sum^{S-1}_{k=0}{\left(
\myvector{y}^{(k)} - 
\sum_{i_0=1}^{I_0} \cdots
\sum_{i_{N}=1}^{I_{N}}{
\sum_{r=0}^{R-1}\myexpectation{}{\lambda_n}
\mymatrix{U}_{i_{0}r_{0}}^{(0)} \dots \mymatrix{U}_{i_{N}r_{N}}^{(N)}
}
(\mytensor{X}^{(k)})^{\star 2}
\right)
}^2 \\
& = \frac{1}{S-1}\sum^{S-1}_{k=0}
\left(\myvector{y}^{(k)} - \theta^{N}\myinner{\mytensor{W}}{\mytensor{X}^{(k)}}
\right)^2
\end{split}
\end{equation*}

\begin{equation*}
\begin{split}
\textrm{V} & =
\frac{1}{S-1}
\sum^{S-1}_{k=0}
{(-1)}^2
\sum_{i_{0}}\cdots\sum_{i_{N}}
{(\mytensor{X}_{i_{0}, \dots,i_{N}}^{(k)})}^2
\sum_{r=0}^{R-1}\myvar{\theta}
(\mymatrix{U}_{i_{0},r}^{(0)})^2 \dots (\mymatrix{U}_{i_{N}r}^{(N)})^2 \\
& =\frac{\theta(1-\theta)}{S-1}
\sum^{S-1}_{k=0}
\myinner{
(\mymatrix{U}^{(0)})^{\star 2}, \dots, \mymatrix{U}^{(N)})^{\star 2}}
{(\mytensor{X}^{(k)})^{\star 2}}
\end{split}
\end{equation*}

Therefore,
\begin{equation}\nonumber
\begin{split}
\mathbb{E}_{ \, \myvector{\lambda}} \bm{ \big[ }\,
    \frac{1}{S-1}\sum^{S-1}_{k=0}
    \left(\myvector{y}^{(k)} - 
        \myinner{\mykruskal{\myvector{\lambda};\, \mathbf{U}^{(0)}, \cdots, \mathbf{U}^{(N)}}}{\mytensor{X}^{(k)}}
    \right)^2
\,\bm{ \big] }
= 
    & \frac{1}{S-1}\sum^{S-1}_{k=0}
    \left(\myvector{y}^{(k)} - 
    \theta\myinner{\mykruskal{\mathbf{U}^{(0)}, \cdots, \mathbf{U}^{(N)}}}{\mytensor{X}^{(k)}}
    \right)^2 \\
    & + \frac{\theta(1-\theta)}{S-1} \sum^{S-1}_{k=0}
    \myinner{\mykruskal{(\mathbf{U}^{(0)})^{\star 2}, \cdots, (\mathbf{U}^{(N)})^{\star 2}}}{(\mytensor{X}^{(k)})^{\star 2}} 
\end{split}
\end{equation}

\end{proof}

\end{document}